\documentclass[11pt,letterpaper]{template}

\usepackage{algorithm}
\usepackage{algpseudocode}
\usepackage{setspace}
\usepackage{etoolbox}
\AtBeginEnvironment{algorithmic}{\setstretch{1.1}}
\usepackage[all]{hypcap}
\usepackage[authoryear,round]{natbib}
\usepackage{mathtools}
\usepackage{multirow}
\usepackage{array}
\usepackage{makecell}
\usepackage[normalem]{ulem}
\usepackage[section]{placeins}

\definecolor{title_blue}{HTML}{204899}
\definecolor{cite_blue}{HTML}{044DC1}
\definecolor{rule_gray}{HTML}{D7DBE2}
\definecolor{rule_blue}{HTML}{4F81BD}
\definecolor{abs_border}{HTML}{DBE4F3}
\definecolor{ours_bg}{HTML}{FDEBD6}
\newcommand{\ourrow}{\rowcolor{ours_bg}}
\newcolumntype{O}{>{\columncolor{ours_bg}}c}
\mdfdefinestyle{absstyle}{
  backgroundcolor=white,
  linecolor=title_blue,
  linewidth=1.0pt,
  roundcorner=10pt,
  innertopmargin=6pt,
  innerbottommargin=12pt,
  innerleftmargin=16pt,
  innerrightmargin=16pt,
  skipabove=0pt,
  skipbelow=0pt,
}
\renewcommand{\titlefont}{\color{title_blue}\normalfont\bfseries\fontsize{14}{17}\selectfont}
\renewcommand{\absfont}{\linespread{1.1}\fontsize{9.5}{12.0}\color{black}\selectfont}
\renewcommand{\HorRule}{\color{rule_gray}\rule{\textwidth}{0.4pt}}
\usepackage{fontawesome5}
\newcommand{\hficon}{\raisebox{-0.15ex}{\includegraphics[height=1.0em]{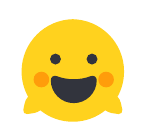}}}

\hypersetup{
  colorlinks=true,
  citecolor=cite_blue,
  linkcolor=magenta,
  urlcolor=magenta
}

\graphicspath{{figures/}}

\newcommand{\hbf}{\mathbf{h}}
\newcommand{\cbf}{\mathbf{c}}
\newcommand{\code}[1]{c_{#1}}
\newcommand{\Kcb}{K}

\renewcommand{\today}{}
\renewcommand{\monthyeardate}[1]{}
\title{{\spaceskip=0.30em\raisebox{-0.40em}{\includegraphics[height=1.5em]{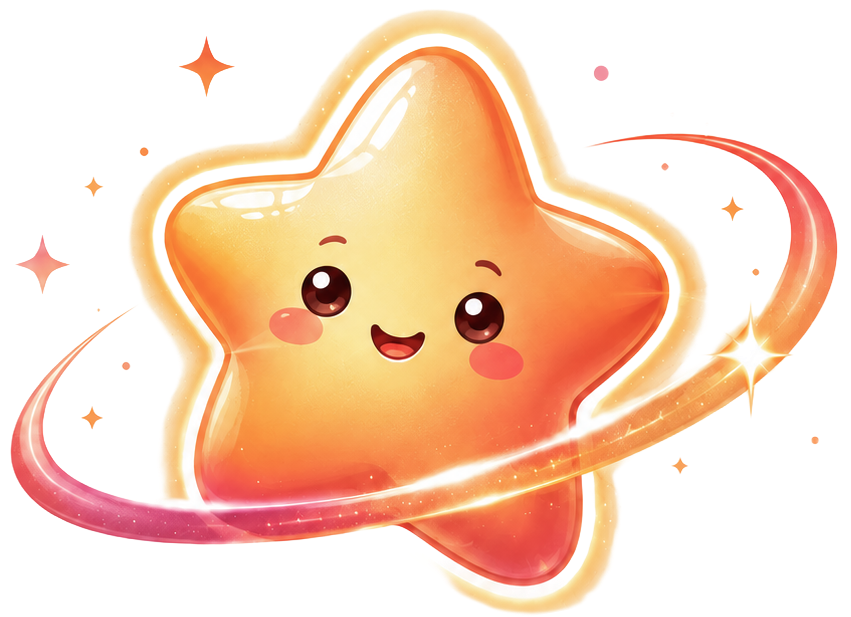}}\hspace{0.15em}Intern Lumina U2: A Multi-Codebook Diffusion Large Language Model for Omni-Visual Understanding and Image Generation}}

\author[1,3*]{Qian~Yu}
\author[1,4*]{Zongyang~Qiu}
\author[1,8*]{Juncheng~Yan}
\author[1*]{Hongliang~Li}
\author[3*]{Qi~Qin}
\author[1*]{Keqi~Wang}
\author[1,11]{Siqi~Luo}
\author[5]{Junkui~Ju}
\author[1]{Yapeng~Mi}
\author[1]{Tiantian~Lu}
\author[1]{Tian~Liu}
\author[1]{Pengyu~Xie}
\author[1]{Chenhao~Ji}
\author[1]{Minghui~Zhang}
\author[1,7]{Yiang~Zhao}
\author[1]{Siyu~Yan}
\author[1]{Haiying~Xu}
\author[1,8]{Yanbo~Wang}
\author[1]{Jianghan~Shen}
\author[1]{Yuandong~Pu}
\author[1]{Jiayi~Lei}
\author[8]{Borui Zhang}
\author[1]{Jinxin~Shi}
\author[1]{Bo~Zhang}
\author[1]{Yuhang~Zang}
\author[1]{Shenglong~Ye}
\author[1]{Lixin~Gu}
\author[1]{Hongjie~Zhang}
\author[1]{Linghan~Liu}
\author[1]{Xinyuan~Chen}
\author[1]{Yuewen~Cao}
\author[7]{Lihe~Zhang}
\author[2,11]{Xiaohong~Liu}
\author[1]{Junjun~He}
\author[5]{Qiangwei~Liu}
\author[1]{Qi~Zhang}
\author[1]{Yu~Cheng}
\author[1,10\dag]{Tianfan~Xue}
\author[4\dag]{Hui~Xiong}
\author[3\dag]{Chang~Xu}
\author[1,2\dag]{Yu~Qiao}
\author[1\dag]{Yihao~Liu}
\author[1,6\dag]{Bin~Fu}
\affil[1]{\mbox{Shanghai AI Laboratory}}
\affil[2]{\mbox{Shanghai Innovation Institute}}
\affil[3]{\mbox{The University of Sydney}}
\affil[4]{\mbox{HKUST(GZ)}}
\affil[5]{\mbox{Huawei Technologies, Shenzhen, China}}
\affil[6]{\mbox{SIAT, CAS}}
\affil[7]{\mbox{Dalian University of Technology}}
\affil[8]{\mbox{Tsinghua University}}
\affil[9]{\mbox{USTC}}
\affil[10]{\mbox{The Chinese University of Hong Kong}}
\affil[11]{\mbox{Shanghai Jiao Tong University}}

\correspondingauthor{\textsuperscript{*}Equal contribution. \quad \textsuperscript{\dag}Corresponding authors.}
\newcommand{\reslink}[3]{\href{#2}{{\color{title_blue}#1}\hspace{0.3em}\textcolor{title_blue}{\textbf{#3}}}}
\projectlinks{%
  \small
  \reslink{\faGlobe}{https://internlm.github.io/InternLumina-U2}{Project Page}\hspace{1.6em}%
  \reslink{\faGithub}{https://github.com/InternLM/InternLumina-U2}{Code}\hspace{1.6em}%
  \reslink{\hficon}{https://huggingface.co/internlm/InternLumina-U2}{Model}%
}

\begin{abstract}
\noindent
{\color{title_blue}\textbf{Abstract---}}Unified multimodal generation covers discrete, continuous, and hybrid paradigms. Discrete approaches with single-codebook visual tokenizers align naturally with text modeling but have limited representational capacity. Continuous and hybrid approaches retain richer visual representations, but they model text and images in heterogeneous spaces with different training objectives,
potentially introducing optimization conflicts between modalities with misaligned representation spaces. These choices expose a central question: how to efficiently model high-capacity visual representations alongside textual content in a shared optimization space within a single backbone. We present \textbf{Intern Lumina U2,} to our knowledge the first diffusion language model to represent multi-codebook visual tokens in a fully discrete space.
Each visual position is represented by a group of codes from complementary codebooks, increasing its representational capacity while preserving compatibility with discrete text modeling.
At the input, the embeddings are concatenated at each position; at the output, a lightweight multi-codebook autoregressive head predicts the codes within each position sequentially, while the diffusion backbone processes all positions in parallel. Built on the unified 4D representation of AToken, Intern Lumina U2 supports image, video, and 3D inputs within a single backbone and achieves leading performance among unified models in visual understanding, image generation, and image editing.
We further build a full-stack training, inference, and evaluation system on both Ascend and NVIDIA platforms to verify the cross-platform effectiveness of our model.
We also improve the inference efficiency through a stability-aware sampling strategy, which combines confidence-plateau-guided unmasking approach with drift-triggered cache refreshing mechanisms, accelerating image generation by $4.45\times$.

\par
\vspace{-0.4em}
{\centering
\includegraphics[width=\linewidth]{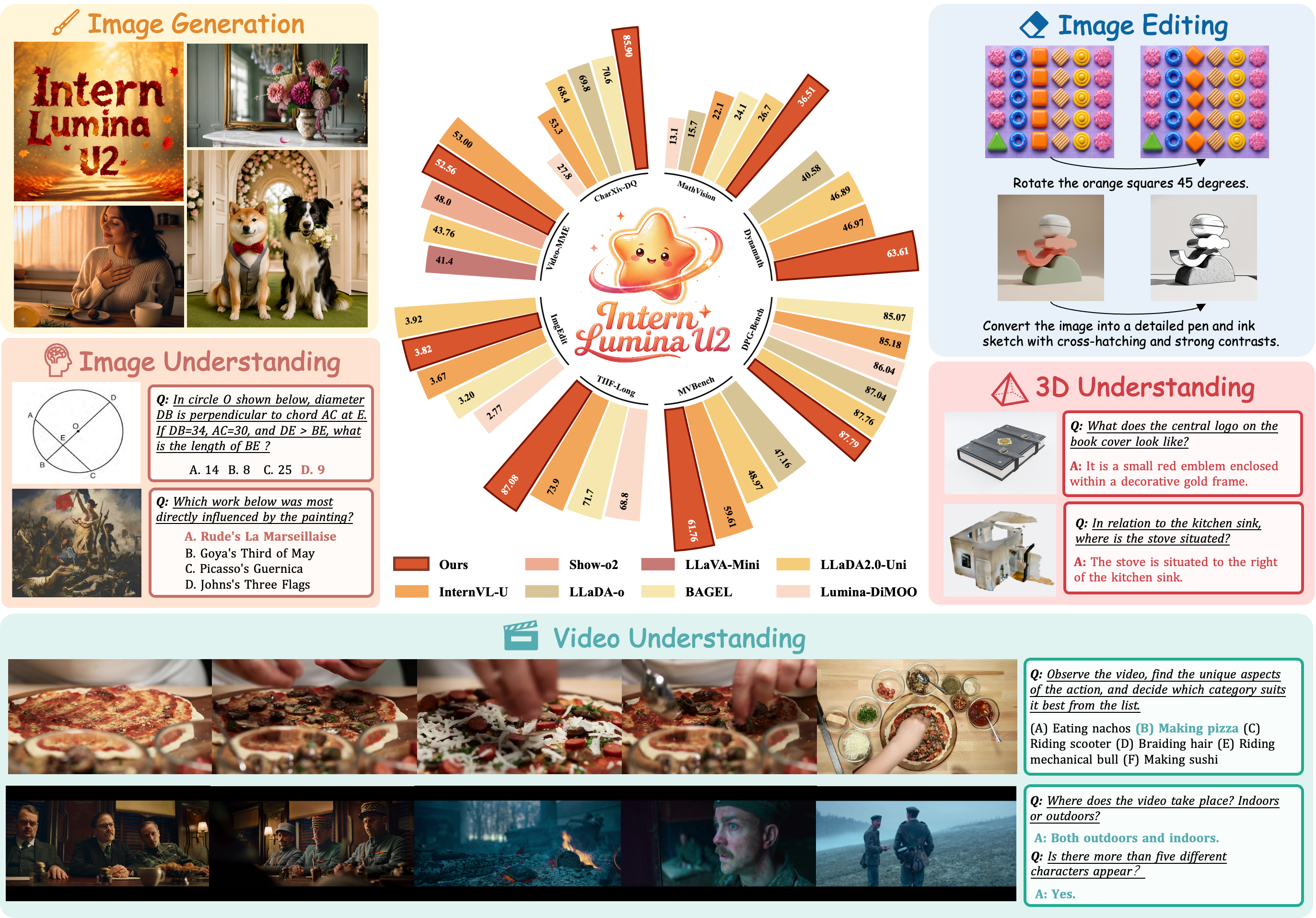}\par
\vspace{-0.5em}
{\absfont\captionof{figure}{Overview of Intern Lumina U2's versatile omni-visual understanding and generation capabilities.}}\par}
\end{abstract}

\begin{document}
\newgeometry{left=2.2cm,right=2.2cm,top=0.0cm,bottom=0.0cm}
\maketitle
\restoregeometry

\clearpage
\thispagestyle{empty}
\begin{center}
  \vspace*{-0.6cm}
  \makebox[\linewidth][c]{%
    \includegraphics[width=1.05\textwidth,height=0.94\textheight,keepaspectratio]{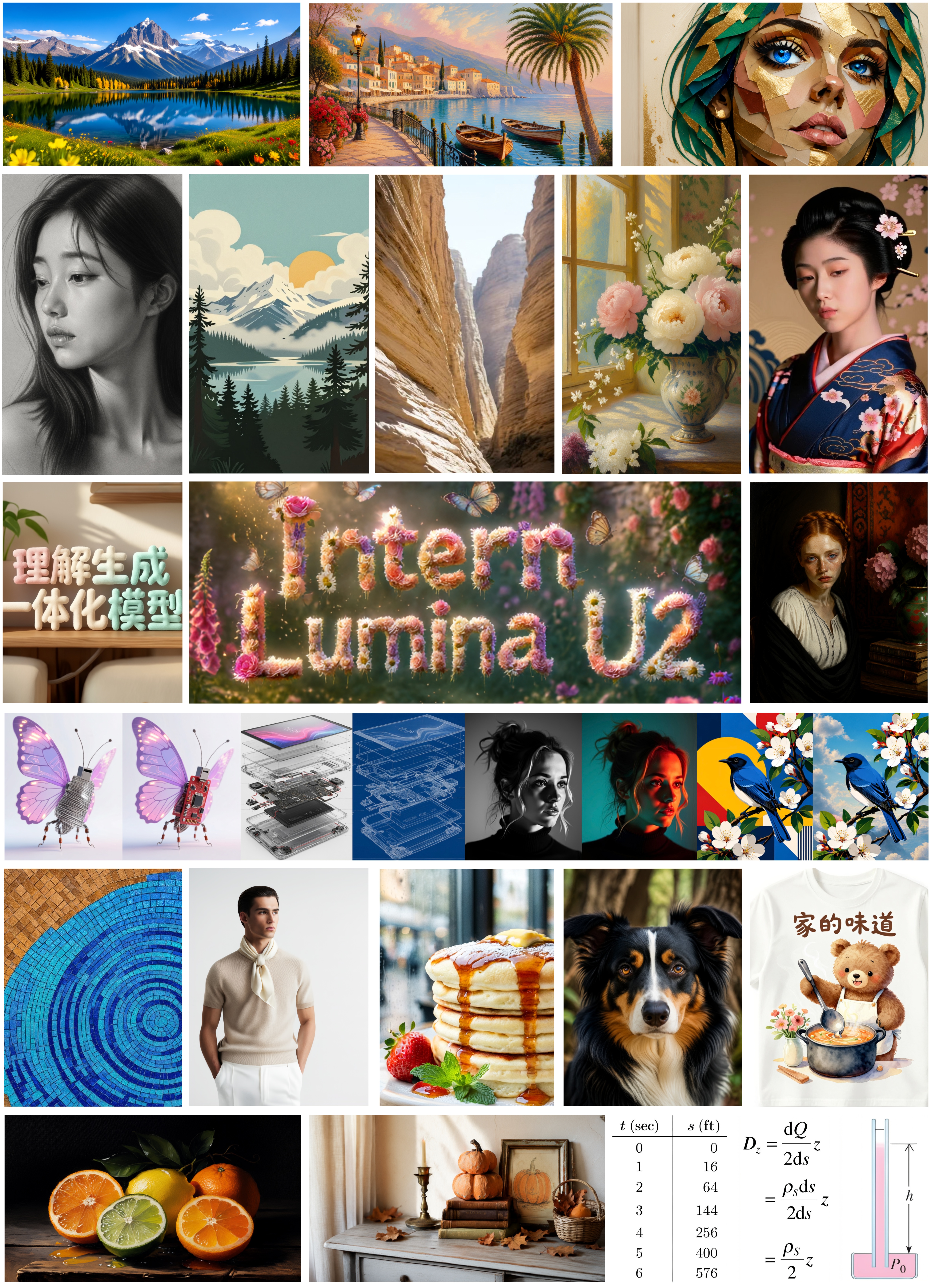}}
  \vspace{-0.2em}
  {\captionsetup{justification=centering}\captionof{figure}{Showcases of Intern Lumina U2 on image generation, editing, and scientific-diagram generation.}}
  \label{fig:effect_gallery}
\end{center}
\clearpage

\section{Introduction}
\label{sec:intro}

Unified multimodal models integrate visual understanding and generation within a single model~\citep{Team2024_Chameleon,Deng2025_BAGEL,Tian2026_InternVLU}.
However, architectural sharing alone does not constitute complete unification, since understanding and generation may still rely on separate representations and training objectives~\citep{Qiu2026_EntryPoint}.
To go beyond architectural unification, we identify two principles for unifying understanding and generation more comprehensively.
First, text and visual content should be mapped into a shared optimization space under a single objective.
Although hybrid designs~\citep{Deng2025_BAGEL,Fan2025_2QRRYMYJ,Tian2026_InternVLU} can achieve high visual fidelity by retaining continuous visual states and delegating synthesis to a continuous decoder, they optimize visual generation with a continuous objective distinct from the text objective and often adopt separate visual encoders for understanding and generation, leading to optimization difficulties.
Second, the generation process should respect the intrinsic structure of each modality.
Text follows a natural sequential order, whereas visual content has no canonical generation order: each region depends on both its local neighborhood and long-range global context.
The generation process should therefore decode text sequentially while refining visual content in parallel with bidirectional context.

Discrete masked diffusion provides a natural basis for satisfying the requirements above. It applies a masked-prediction objective to discrete text and visual tokens, thereby optimizing both modalities within a common discrete space.
Under this formulation, each task is specified by partitioning the token sequence into observed conditioning tokens and masked prediction targets. Understanding and generation are thereby cast as distinct conditional queries to the same model, rather than as separate tasks with task-specific objectives.
Moreover, masked diffusion decodes tokens in parallel through iterative refinement without imposing a fixed generation order~\citep{Nie2025_LLaDA,Chang2022_MaskGIT,Yang2025_6PQB9975,Xin2025_26M7KSHA}, allowing each spatial position to be refined with bidirectional access to both local and global context.
However, existing discrete unified models~\citep{Xin2025_26M7KSHA,Xie2025_FEMHL63A,Yang2025_6PQB9975} remain limited in visual representation capacity, as single-codebook tokenization compresses the visual information at each position into a single discrete code. Such a compact representation is insufficient both for faithful reconstruction and for covering the full range of visual tasks~\citep{Ma2025_UniTok}.
This motivates us to design a unified discrete framework that carries a high-capacity visual representation in a shared space under a single training objective, while modeling it efficiently with a unified backbone.

To this end, we present Intern Lumina U2, to our knowledge the first diffusion language model that models multi-codebook visual states in a fully discrete space, with AToken~\citep{Lu2025_XVFGVESC} as the visual tokenizer and LLaDA2.0~\citep{llada2} as the backbone. Intern Lumina U2 encodes images, videos, and 3D assets into a unified 4D code space and organizes these visual tokens together with text tokens in a unified packed-sequence format, so that understanding and generation across all these modalities are formulated over the same sequence representation and handled by a shared backbone. Each visual position carries $\Kcb$ complementary codes: independent embedding tables concatenated at the input preserve codebook identity for efficient training, while a lightweight multi-codebook autoregressive head models the dependencies among the codes within each position. Within this shared representation, a multi-task masking mechanism assigns each task its own masking scheme, so that understanding, generation, and editing across modalities are trained jointly under a single objective. For image generation, a stability-aware inference strategy commits tokens at confidence plateaus and refreshes the cache on hidden-state drift, achieving a $4.45\times$ end-to-end speedup. We further build the operators, parallelization strategies, and system adaptations for training, inference, and evaluation on 1,024 Ascend 910B NPUs, with a smaller-scale NVIDIA reference run using the same architecture and optimization recipe but a smaller training budget for cross-platform validation.

Extensive experiments on image, video, and 3D understanding, as well as image generation and editing, demonstrate the effectiveness of this unified design.
To our knowledge, Intern Lumina U2 is the first diffusion language model to support this full combination of visual modalities and capabilities.
Trained on data curated by a unified pipeline that organizes heterogeneous sources under a common protocol, the model achieves leading performance among unified multimodal models across our evaluations. With only 1.4B activated parameters, it ranks first on six of ten image-understanding benchmarks and improves on eight of ten over the same-backbone LLaDA2.0-Uni~\citep{Bie2026_LLaDA2Uni}.
It also shows superior performance on all three video benchmarks~\citep{Fu2024_VideoMME,Zhou2024_MLVU,Li2024_MVBench} and on 3D~MM-Vet~\citep{Qi2024_ShapeLLM}, and among unified models it delivers strong text-to-image generation on DPG-Bench~\citep{Hu2024_DPG}, UniGenBench~\citep{Wang2025_UniGenBench}, TIIF-Bench~\citep{Wei2025_TIIF}, and GenEval~\citep{Ghosh2023_GenEval}. It also performs strongly on physics-faithful scientific-diagram generation. These results establish Intern Lumina U2 as a competitive unified discrete foundation model for omni-visual understanding and generation.

\paragraph{Contributions.}
\begin{itemize}[leftmargin=1.4em,itemsep=2pt,topsep=2pt]
    \item \textbf{Multi-codebook discrete modeling for unified dLLMs.}
    To our knowledge, Intern Lumina U2 is the first unified diffusion language model built on multi-codebook visual states, with visual prediction factorized across spatial positions and codebooks.

    \item \textbf{Omni-visual understanding in a unified discrete space.}
    Intern Lumina U2 is the first diffusion language model to unify visual understanding across images, videos, and 3D assets with image generation and editing, moving unified multimodal modeling toward a general omni-visual paradigm.

    \item \textbf{Full-stack adaptation and validation on Ascend.}
    We complete the full-stack adaptation of Intern Lumina U2 on 1,024 Ascend 910B NPUs, enabling end-to-end training, inference, and comprehensive evaluation on the platform.

    \item \textbf{Efficient image-generation inference.}
    We introduce Plateau-Halton sampling with V-verify (PHV), which commits tokens at confidence plateaus and refreshes the cache on hidden-state drift, achieving a $4.45\times$ speedup in end-to-end image-generation latency with a favorable quality--efficiency trade-off.

    \item \textbf{Broad empirical validation.}
    Intern Lumina U2 achieves leading performance across image, video, and 3D understanding, text-to-image generation, image editing, and scientific-diagram generation, demonstrating the effectiveness and breadth of the proposed unified framework.
\end{itemize}

\section{Related Work}
\label{sec:related}

We relate Intern Lumina U2 to four bodies of work: unified multimodal models, diffusion language models, multi-codebook visual representations, and inference acceleration for masked diffusion.
Table~\ref{tab:related_axes} groups representative systems by backbone paradigm and summarizes their capability and design profile; Intern Lumina U2 is the only system in the comparison that combines image generation, perception across images, videos, and 3D assets, and masked-diffusion decoding over a multi-codebook visual state. It represents all three visual input modalities with discrete backbone tokens and generates images without a separate continuous diffusion or flow generator.

\paragraph{Unified multimodal understanding and generation.}
One line of work models image generation as discrete visual-token prediction within the backbone. Chameleon and Emu3 use a shared vocabulary for text and visual codes under next-token prediction~\citep{Team2024_Chameleon,Wang2024_Emu3}, whereas MMaDA and Lumina-DiMOO use masked prediction for both modalities~\citep{Yang2025_6PQB9975,Xin2025_26M7KSHA}. Related discrete-generation designs include Janus and Janus-Pro, which decouple the understanding encoder from the generation tokenizer~\citep{Wu2024_KRKE6NRZ,Chen2025_JanusPro}; VAR and OneCAT, which use coarse-to-fine next-scale prediction~\citep{Tian2024_VAR,Li2025_OneCAT}; and Emu3.5, which converts next-token image decoding to bidirectional parallel prediction for faster inference~\citep{Cui2025_Emu35}.
A second line models image generation in a continuous space with a diffusion or flow objective. Transfusion combines text next-token loss with image diffusion in one Transformer~\citep{Zhou2024_N4DD8N5Z}, UniFluid models each position with a diffusion head~\citep{Fan2025_2QRRYMYJ}, BAGEL and InternVL-U use ViT features for understanding and VAE-space flow matching for generation~\citep{Deng2025_BAGEL,Tian2026_InternVLU}, and SenseNova-U1 moves flow matching to pixel space~\citep{SenseNova2026_U1}.
A third, hybrid route predicts discrete semantic visual tokens in the backbone and reconstructs images with a separate continuous generator. LLaDA2.0-Uni follows this design: its backbone models text and visual tokens under a shared masked-prediction objective, while a diffusion decoder synthesizes images conditioned on the predicted visual tokens~\citep{Bie2026_LLaDA2Uni}.
Single-codebook discrete representations constrain the detail and semantics carried at each visual position. Continuous and hybrid designs support richer image synthesis, but retain a continuous generative objective distinct from text-token classification, either within the backbone or in a separate generator.
Intern Lumina U2 instead uses multi-codebook discrete representations for image, video, and 3D inputs, while training text prediction and image generation under a shared objective.

\paragraph{Diffusion language models.}
LLaDA showed that masked diffusion language models trained from scratch can match strong AR baselines, replacing left-to-right decoding with parallel iterative recovery~\citep{Nie2025_LLaDA}, and LLaDA2.0 converts a pretrained AR Mixture-of-Experts model into a block-wise diffusion LM, preserving its knowledge while enabling parallel decoding~\citep{llada2}.
This family has been extended to vision in several ways: LLaDA-V feeds features from a vision encoder into the diffusion LM for understanding~\citep{You2025_LLaDAV}, and LLaDA-o pairs discrete masked diffusion for text with continuous diffusion for images~\citep{You2026_LLaDAo}. Among the unified models discussed above, MMaDA and Lumina-DiMOO extend masked prediction to both text and image tokens~\citep{Yang2025_6PQB9975,Xin2025_26M7KSHA}, while LLaDA2.0-Uni adopts the converted LLaDA2.0 backbone with block-wise masked prediction~\citep{Bie2026_LLaDA2Uni}.
Intern Lumina U2 also starts from LLaDA2.0-mini and extends this family along two axes: the visual state becomes a multi-codebook visual token, and the same model covers image, video, and 3D inputs.

\paragraph{Multi-codebook representations and depth decoding.}
Residual and factorized quantization raise the capacity of discrete visual states without lengthening the spatial grid: RQ-Transformer decodes residual codes with a spatial-plus-depth Transformer~\citep{Lee2022_2BQBM5BT}, VILA-U brings multi-codebook depth decoding into an AR unified model~\citep{Wu2024_VILAU}, UniTok folds $H{\times}W{\times}\Kcb$ codes into $H{\times}W$ backbone positions~\citep{Ma2025_UniTok}, and TokenBridge documents the dependence that is lost when the codes of one position are predicted independently~\citep{Wang2025_B5RELEDK}.
AToken supplies the shared image/video/3D multi-codebook representation used here~\citep{Lu2025_XVFGVESC}.
These works establish multi-codebook modeling in autoregressive systems. Intern Lumina U2 instead makes the multi-codebook visual token the visual state of a masked-diffusion language model: the backbone refines spatial positions in parallel, while the multi-codebook AR head models the dependence among codes within each position (Section~\ref{sec:mcio}). This factorization keeps the backbone sequence length independent of $\Kcb$ without treating the codebooks as conditionally independent.

\paragraph{Accelerating masked-diffusion decoding.}
Acceleration methods for masked generation follow two routes.
Learning-based methods train an auxiliary model so that few large jumps replace many refinement steps; a recent representative learns latent controlled dynamics for masked image generation~\citep{migm-shortcut}.
Training-free methods instead reshape the sampler: MaskGIT-style schedules fix how many positions commit per round~\citep{Chang2022_MaskGIT}, Fast-dLLM commits every position whose probability clears an absolute threshold~\citep{Wu2025_FastdLLM}, low-discrepancy reveal orders improve which positions commit~\citep{Besnier2025_Halton}, and cached sampler state with fixed-interval refresh removes redundant backbone work~\citep{Xin2025_26M7KSHA}.
Our sampler follows the training-free route. Because visual-token confidence sits far below the absolute thresholds of Fast-dLLM-style commit rules (Section~\ref{sec:confidence_stability}), it adapts both the commit schedule and the cache refresh from temporal signals, confidence plateaus and measured hidden-state drift (Section~\ref{sec:inference_accel}).

\newcommand{\cmark}{\ding{51}}
\newcommand{\xmark}{\textcolor{black!30}{\ding{55}}}
\begin{table}[!t]
  \centering
  \small
  \renewcommand{\arraystretch}{1.15}
  \caption{Capability and design checklist for representative unified multimodal models, grouped by backbone paradigm.
  \emph{Discrete tokens} refers to the visual state inside the backbone; \emph{Masked-diff. decoding} marks models that generate visual tokens by iterative masked denoising rather than left-to-right prediction; \emph{Multi-codebook} denotes multiple codes per visual position with their dependence modeled explicitly; \emph{Diffusion text} marks models that also generate text through the masked-diffusion process rather than autoregressively; \emph{No continuous generator} marks systems that decode images through backbone tokens, without a separate continuous (diffusion or flow) generator.
  Systems: Emu3~\citep{Wang2024_Emu3}, Emu3.5~\citep{Cui2025_Emu35}, Janus-Pro~\citep{Chen2025_JanusPro}, VILA-U~\citep{Wu2024_VILAU}, UniTok~\citep{Ma2025_UniTok}, BAGEL~\citep{Deng2025_BAGEL}, InternVL-U~\citep{Tian2026_InternVLU}, SenseNova-U1~\citep{SenseNova2026_U1}, Show-o2~\citep{Xie2025_Showo2}, Show-o~\citep{Xie2025_FEMHL63A}, MMaDA~\citep{Yang2025_6PQB9975}, Lavida-O~\citep{Li2025_LavidaO}, Lumina-DiMOO~\citep{Xin2025_26M7KSHA}, LLaDA2.0-Uni~\citep{Bie2026_LLaDA2Uni}.}
  \label{tab:related_axes}
  \setlength{\tabcolsep}{4.5pt}
  \resizebox{\linewidth}{!}{%
  \begin{tabular}{lccccccccc}
    \toprule
    & \multicolumn{4}{c}{Supported modalities} & \multicolumn{5}{c}{Representation and decoding} \\
    \cmidrule(lr){2-5}\cmidrule(lr){6-10}
    Model & \makecell{Image\\und.} & \makecell{Image\\gen.} & \makecell{Video\\und.} & \makecell{3D\\und.} & \makecell{Discrete\\tokens} & \makecell{Masked-diff.\\decoding} & \makecell{Multi-\\codebook} & \makecell{Diffusion\\text} & \makecell{No continuous\\generator} \\
    \midrule
    \multicolumn{10}{l}{\emph{Autoregressive}} \\
    Emu3 & \cmark & \cmark & \cmark & \xmark & \cmark & \xmark & \xmark & \xmark & \cmark \\
    Emu3.5 & \cmark & \cmark & \xmark & \xmark & \cmark & \xmark & \xmark & \xmark & \cmark \\
    Janus-Pro & \cmark & \cmark & \xmark & \xmark & \cmark & \xmark & \xmark & \xmark & \cmark \\
    VILA-U & \cmark & \cmark & \cmark & \xmark & \cmark & \xmark & \cmark & \xmark & \cmark \\
    UniTok & \cmark & \cmark & \xmark & \xmark & \cmark & \xmark & \cmark & \xmark & \cmark \\
    BAGEL & \cmark & \cmark & \xmark & \xmark & \xmark & \xmark & \xmark & \xmark & \xmark \\
    InternVL-U & \cmark & \cmark & \xmark & \xmark & \xmark & \xmark & \xmark & \xmark & \xmark \\
    SenseNova-U1 & \cmark & \cmark & \xmark & \xmark & \xmark & \xmark & \xmark & \xmark & \xmark \\
    Show-o2 & \cmark & \cmark & \cmark & \xmark & \xmark & \xmark & \xmark & \xmark & \xmark \\
    \addlinespace
    \multicolumn{10}{l}{\emph{Masked diffusion}} \\
    Show-o & \cmark & \cmark & \xmark & \xmark & \cmark & \cmark & \xmark & \xmark & \cmark \\
    MMaDA & \cmark & \cmark & \xmark & \xmark & \cmark & \cmark & \xmark & \cmark & \cmark \\
    Lavida-O & \cmark & \cmark & \xmark & \xmark & \cmark & \cmark & \xmark & \cmark & \cmark \\
    Lumina-DiMOO & \cmark & \cmark & \xmark & \xmark & \cmark & \cmark & \xmark & \cmark & \cmark \\
    LLaDA2.0-Uni & \cmark & \cmark & \xmark & \xmark & \cmark & \cmark & \xmark & \cmark & \xmark \\
    \rowcolor{ours_bg}
    \textbf{Intern Lumina U2} & \cmark & \cmark & \cmark & \cmark & \cmark & \cmark & \cmark & \cmark & \cmark \\
    \bottomrule
  \end{tabular}}
\end{table}

\section{Model Architecture}
\label{sec:arch}

\subsection{Overview}
\label{sec:arch_overview}

\begin{figure}[!t]
  \centering
  \includegraphics[width=\linewidth]{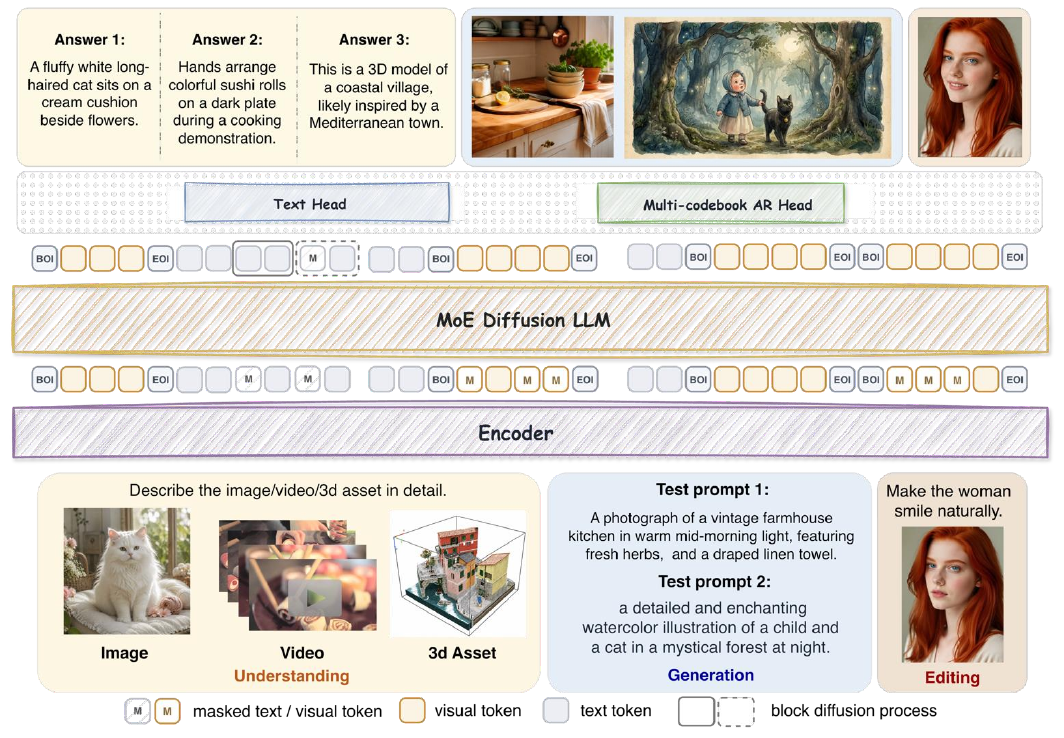}
  \caption{Overall architecture of Intern Lumina U2. Visual inputs are encoded into visual tokens by AToken and text into text tokens; the MoE diffusion LLM recovers masked text tokens through the text head, with text answers generated by block diffusion, and masked visual tokens through the multi-codebook AR head, which predicts the $\Kcb$ codes of each visual token sequentially.}
  \label{fig:arch}
\end{figure}

As shown in Figure~\ref{fig:arch}, Intern Lumina U2 is a masked diffusion language model that supports visual understanding over images, videos, and 3D assets, as well as image generation and editing within a unified model (Figure~\ref{fig:arch}).
To model all tasks with a single backbone and a shared training objective, we represent every modality as discrete tokens.
Specifically, visual inputs (images, videos, and 3D assets) are first converted into visual tokens by a frozen visual tokenizer, AToken (Section~\ref{sec:tokenizer}), while text is converted into text tokens by the native tokenizer of the base language model, LLaDA2.0-mini.
The resulting visual and text tokens, together with the target tokens (either a text answer or a target image, represented as the mask tokens), are arranged into a single sequence.
The MoE diffusion LLM (Section~\ref{sec:backbone}) processes this sequence and predicts the masked tokens in a  modality-specific manner.
Masked text tokens are predicted by a text head and generated block by block.
In contrast, masked visual tokens are predicted by a multi-codebook AR head (Section~\ref{sec:mcio}): each visual token contains $\Kcb{=}8$ discrete codes, and the head predicts visual tokens at different positions in parallel while generating the discrete codes within each token sequentially.
Finally, the AToken decoder maps the predicted visual codes back to pixels.

\subsection{Unified Visual Tokenizer}
\label{sec:tokenizer}

The visual tokenizer serves as the interface between raw visual data (images, videos, and 3D assets) and the model, and therefore constrains both generation and understanding.
As discussed in the previous section, Intern Lumina U2 places two requirements on its visual tokenizer: (1) it must produce discrete visual tokens that can be modeled in the same way as text tokens, so that all tasks can share a single masked-diffusion objective; and (2) these tokens must preserve both the low-level detail for generation and the high-level semantics needed for understanding.
However, existing tokenizers are typically specialized for either low-level detail or high-level semantics.
For example, tokenizers trained for reconstruction, such as VQGAN~\citep{Esser2021_VQGAN}, keep fine detail but are not trained to align with language, whereas semantic encoders such as SigLIP~\citep{Zhai2023_SigLIP} align well with text but are not trained to reconstruct pixels.

We therefore adopt AToken~\citep{Lu2025_XVFGVESC} as the visual tokenizer. AToken is initialized from a pretrained SigLIP2 encoder~\citep{Tschannen2025_SigLIP2} and trained jointly for reconstruction and understanding. As a result, its tokens carry semantic structure while still supporting faithful reconstruction. In addition, AToken encodes images, videos, and 3D assets into a shared 4D latent space, so that a single tokenizer serves all three modalities. We use its discrete configuration, which quantizes each location of the latent grid with finite scalar quantization (FSQ)~\citep{Mentzer2023_FSQ} over $\Kcb{=}8$ codebooks. Each codebook is a vocabulary of $4096$ entries, and a \emph{code} is the index of one entry, so each location is represented by $\Kcb$ codes $\cbf=(\code{1},\dots,\code{\Kcb})$, one per codebook. We call such a location a \emph{visual position} and its $\Kcb$ codes a \emph{visual token} (Figure~\ref{fig:arch}). Compared with a single code per position, the eight codes give the representation more capacity for fine visual detail. The tokenizer is kept frozen during training, and the grid shapes, flattening orders, and structural tokens are listed in Appendix~\ref{app:layout}.

\subsection{Backbone}
\label{sec:backbone}

We build our backbone upon LLaDA2.0-mini~\citep{llada2}, a masked diffusion language model obtained by converting a pretrained autoregressive mixture-of-experts (MoE) language model~\citep{LingTeam2025_Ling2} into a block diffusion model.
It has $16$B parameters in total, of which $1.4$B are activated per token.
Among its 20 Transformer layers, the first employs a dense feed-forward network, while the remaining 19 are MoE layers.
Each MoE layer consists of 256 routed experts and a single shared expert, with 8 routed experts activated per token.
LLaDA2.0-mini is a language model which generates text via block diffusion: a sequence is divided into blocks of consecutive tokens, and each block is denoised by masked diffusion conditioned on all preceding blocks.
Accordingly, it generates text block by block at inference: the masked tokens within the current block are predicted in parallel and iteratively refined over several denoising steps, while the key--value cache of previously completed blocks is reused.
Since LLaDA2.0-mini already performs masked diffusion over text, we can directly inherit its language capabilities without an additional autoregressive-to-diffusion conversion~\citep{Gong2025_DiffuLLaMA,Ye2025_Dream7B,llada2}.

We introduce two modifications to LLaDA2.0-mini while keeping all other components unchanged, including the Transformer layers, the text embedding table, the text head, and the \texttt{<|mask|>} token.
First, we add the visual input and output interfaces described in Section~\ref{sec:mcio}, which account for 0.65B parameters.
Second, we replace the attention mask with our proposed task-aware attention mask.
LLaDA2.0-mini was trained exclusively on text with a block-wise attention mask, whereas our Intern Lumina U2 also contains visual conditions and visual answers.
We therefore design a task-aware attention mask that controls attention among these segments according to the task (Section~\ref{sec:seq_mask}).
It is worthy to mention that we retain the native positional encoding (e.g. 1D RoPE~\citep{Su2021_RoFormer}) of the backbone for visual tokens, rather than introducing a dedicated 2D or 3D scheme, which preserves the positional prior learned in pretraining.
Since 1D RoPE does not explicitly encode spatial structure, the spatial or spatio-temporal layout of each visual input is instead indicated by structural delimiter tokens inserted into the sequence (Appendix~\ref{app:layout}).
In this way, images, videos, and 3D assets are all represented in the same 1D token sequence as text.

\subsection{Multi-Codebook Input and Output}
\label{sec:mcio}

With a multi-codebook tokenizer, each visual position is represented by $\Kcb$ codes, all of which must be encoded at the input and predicted at the output. We compare three designs for handling the $\Kcb$ codes of a visual token. The first flattens them into $\Kcb$ separate tokens, which multiplies the sequence length by $\Kcb$. The second keeps one backbone token per visual position but predicts the $\Kcb$ codes independently, which ignores the dependence among them. The third keeps one backbone token per visual position and models the $\Kcb$ codes jointly. Our Intern Lumina U2 adopts the third design: at the input, the codes of each visual position are merged into a single backbone token, and at the output, they are predicted jointly by the multi-codebook AR head.
In the remainder of this section, we provide the detailed design of our multi-codebook input and output interface. We use $\cbf=(\code{1},\dots,\code{\Kcb})$ to denote the $\Kcb$ codes at a visual position, $e_k=E_k(\code{k})$ to denote the embedding of code $\code{k}$ (defined below), and $\hbf\in\mathbb{R}^{d}$ ($d{=}2048$) to denote the last-layer hidden state of the backbone at the same position.

\paragraph{Input interface.}
\begin{figure}[!htbp]
  \centering
  \includegraphics[width=0.98\linewidth]{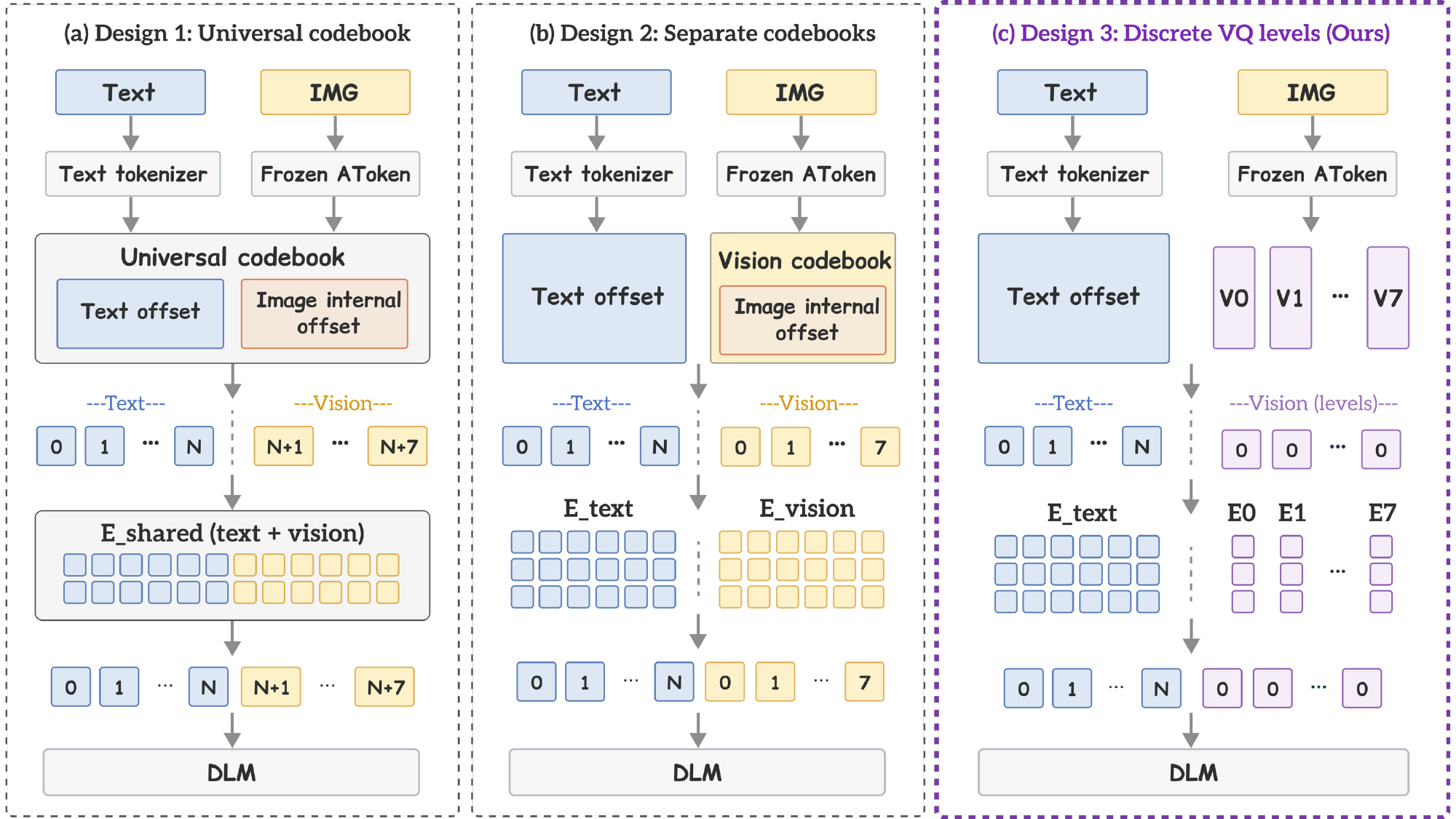}
  \caption{Three candidate input interfaces for multi-codebook visual tokens. We adopt the third design: independent embeddings for the eight codebooks, concatenation at each spatial position, and projection into one backbone token.}
  \label{fig:input_designs}
\end{figure}

The visual input interface maps the $\Kcb$ codes of each visual position to a single vector in the hidden space.
Each codebook $k$ has its own embedding table $E_k$, which holds an embedding for each of its 4096 codes, together with a mask entry and several special entries that mark the layout of visual inputs (Table~\ref{tab:tokens}).
When a visual position is masked, all of its $\Kcb$ codes are replaced with the mask entry.
The input vector $\mathbf{x}$ of a visual position is obtained by concatenating its $\Kcb$ code embeddings and applying a linear projection,
\begin{equation}
  \mathbf{x} \;=\; W_{\mathrm{in}}\,
  \big[\,E_1(\code{1});\,E_2(\code{2});\,\dots;\,E_{\Kcb}(\code{\Kcb})\,\big],
  \qquad W_{\mathrm{in}}\in\mathbb{R}^{d\times \Kcb d}.
  \label{eq:uninput}
\end{equation}
This design offers two advantages.
First, merging the codes of each position before they enter the backbone keeps the sequence length independent of $\Kcb$.
Second, since each codebook has its own embedding table, the visual codes need not be added to the text vocabulary, and no index offsets are needed to distinguish the codebooks.
To demonstrate the effectiveness of our design, we compare it with two alternatives that use shared embedding tables (in Table~\ref{tab:input_ablation} and Figure~\ref{fig:input_ablation}).

\paragraph{Output interface: multi-codebook AR head.}

The visual output interface predicts the $\Kcb$ codes of each visual position from the single hidden state $\hbf$.
Similar to the depth transformers of RQ-Transformer~\citep{Lee2022_2BQBM5BT} and VILA-U~\citep{Wu2024_VILAU}, we capture the dependencies among these codes by factorizing their joint distribution autoregressively along the codebook index:
\begin{equation}
  p(\cbf \mid \hbf) \;=\; \prod_{k=1}^{\Kcb}
  p\!\left(\code{k} \,\middle|\, \hbf,\, c_{<k}\right).
  \label{eq:depthar}
\end{equation}
We implement this factorization with a \emph{multi-codebook AR head}, a 6-layer causal Transformer of width $d$ that performs the prediction autoregressively along the codebook dimension.
For each visual position, the input to the head is a sequence of length $\Kcb$, which we refer to as the \emph{depth sequence}.
The first element is $q+s_1$, where $q=W_q\,\hbf+b_q$ is a linear projection of the backbone hidden state with bias $b_q$; for $k\ge 2$, the $k$-th element is $e_{k-1}+s_k$, where $e_{k-1}=E_{k-1}(\code{k-1})$ is the embedding of the preceding code, obtained from the embedding tables of the input interface (Eq.~\eqref{eq:uninput}), and $s_k$ is a learnable embedding of position $k$ in the depth sequence.
In addition, each block of the head is conditioned through adaptive LayerNorm (adaLN)~\citep{Peebles2023_DiT} on $q+u_k$ at position $k$, where $u_k$ is a second learnable embedding of position $k$, so that the backbone output directly modulates every prediction step.
Finally, a separate classifier $W^{\mathrm{out}}_k\in\mathbb{R}^{4096\times d}$ for each codebook maps the output of the head at position $k$ to the logits for $\code{k}$.

To validate this factorization, we compare the multi-codebook AR head with a parallel head that predicts the codes independently, $p_{\mathrm{par}}(\cbf\mid\hbf)=\prod_k p(\code{k}\mid\hbf)$.
Note that the parallel head is a special case of Eq.~\eqref{eq:depthar} in which each factor drops the conditioning on $c_{<k}$, while Eq.~\eqref{eq:depthar} follows directly from the chain rule and makes no independence assumption.
Consequently, with ideal conditional models, the cross-entropy of the multi-codebook AR head is never higher than that of the parallel head, with equality if and only if the $\Kcb$ codes are conditionally independent given $\hbf$.
Table~\ref{tab:head_ablation} quantifies this gain in practice.

\begin{figure}[!htbp]
  \centering
  \includegraphics[width=\linewidth]{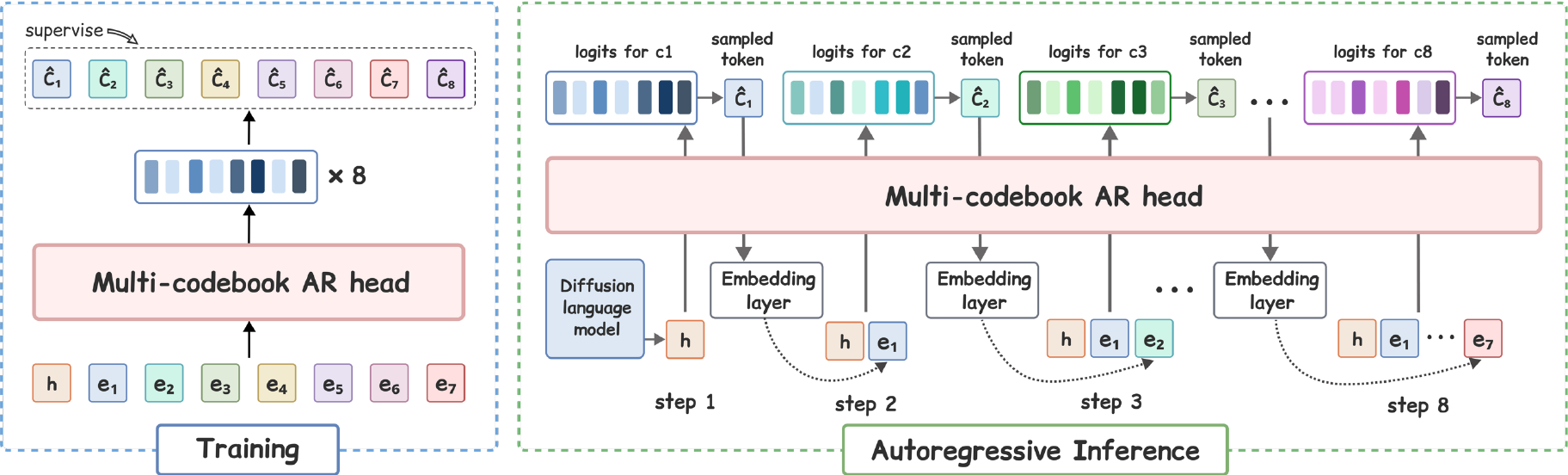}
  \caption{Training and inference of the multi-codebook AR head: one teacher-forced causal pass yields all $\Kcb$ logits at training; at inference the prefix grows for $\Kcb$ steps, each sampled code $\hat c_k$ is embedded into $e_k$ and appended. Positions are processed in parallel, codebooks autoregressively.}
  \label{fig:archead}
\end{figure}

During training, our multi-codebook AR head is optimized with teacher forcing (Figure~\ref{fig:archead}). For each masked visual position, the depth sequence is constructed from the ground-truth codes:
\begin{equation}
  \big[\,q+s_1,\; e_1^{\mathrm{gt}} + s_2,\; \dots,\;
  e_{\Kcb-1}^{\mathrm{gt}} + s_{\Kcb}\,\big],
\end{equation}
where $e_k^{\mathrm{gt}}=E_k(\code{k}^{\mathrm{gt}})$ is the embedding of the ground-truth code $\code{k}^{\mathrm{gt}}$ of codebook $k$. Because the head is causal, a single forward pass produces the logits of all $\Kcb$ codes, with the $k$-th output predicting $\code{k}$.
At inference, the ground-truth codes are not available. The multi-codebook AR head therefore predicts the codes of a position in $\Kcb$ sequential steps: at step $k$, it samples $\code{k}$ and, when $k<\Kcb$, appends its embedding $e_k+s_{k+1}$ to the depth sequence for step $k+1$.
Different masked positions are processed in parallel throughout.
Section~\ref{sec:inference} details this sampling procedure and its caching strategy.

\subsection{Unified Data Format with Data Packing}
\label{sec:seq_mask}

Having specified the representation of a single visual position, we now describe the organization of complete samples into training sequences.
All tasks, from text-only dialogue to image generation and editing, share a unified packed sequence format.
Moreover, to determine the attention mask and calculate loss, each token in this format is annotated with five metadata fields: \emph{type} (text condition, visual condition, structural token, text answer, or visual answer), \emph{turn index}, \emph{noise state} (clean or noisy), \emph{block index}, and \emph{position ID}.

\paragraph{Dialogue structure and sequence layout.}
Every sample is first formatted as a dialogue between a user and an assistant, in which each turn consists of a user message and the assistant's reply to it.
The \emph{conditions}, provided by the user, consist of a text prompt, an image, a video, a 3D asset, or any combination of them.
The \emph{answers}, provided by the assistant, are partly masked during training and serve as the prediction targets.
Text answers and visual answers follow different layouts, which we describe in the following.

\paragraph{Text answers.}
A text answer consists of \texttt{<BOA>}, the answer tokens, and \texttt{<EOA>}, and its final block is padded to a multiple of the block length $B$ with additional \texttt{<EOA>} tokens \emph{before} corruption.
End-of-answer padding thus becomes an ordinary supervised target, and every block contains exactly $B$ maskable tokens.

\begin{figure}[!htbp]
  \centering
  \includegraphics[width=\linewidth]{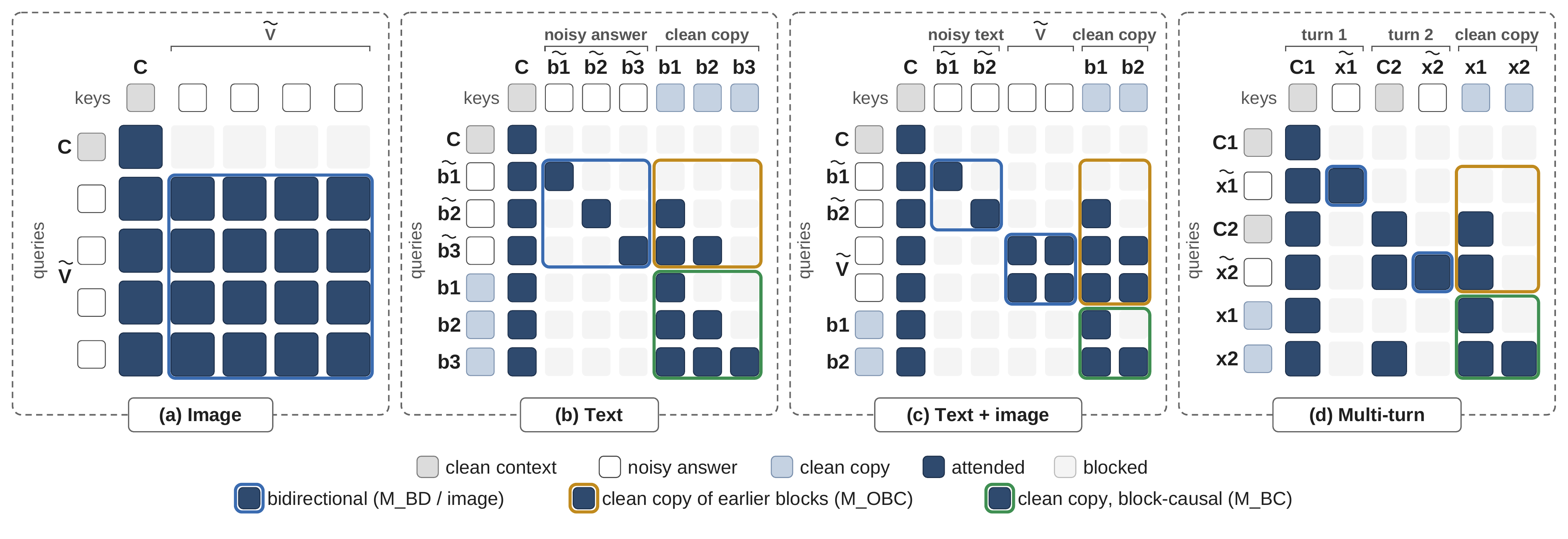}
  \caption{Task-aware attention mask at block level. (a)~Image answer: one noisy visual answer $\tilde V$ after the clean context $C$, bidirectional over all positions. (b)~Text answer: block diffusion over the noisy blocks $\tilde b_1,\tilde b_2,\tilde b_3$ and their clean copies $b_1,b_2,b_3$, with $M_{\mathrm{BD}}$, $M_{\mathrm{OBC}}$, and $M_{\mathrm{BC}}$. (c)~A text answer followed by an image answer, as in thinking before generation: the image reads the clean copy of the preceding text blocks. (d)~Multi-turn text with conditions $C_1,C_2$, noisy answers $\tilde x_1,\tilde x_2$, and clean copies $x_1,x_2$: a later turn reads the clean copies of earlier answers.}
  \label{fig:mask}
\end{figure}

When a sample is packed into a sequence, its conditions and \emph{noisy} answers are placed in dialogue order, followed by a \emph{clean copy} of each text answer:
\begin{equation*}
  [\;\underbrace{\text{clean context} + \text{noisy answer}}_{\text{dialogue stream}}
  \;\mid\;
  \underbrace{\text{clean copy}}_{\text{clean stream}}\;].
\end{equation*}
The clean copies serve as unmasked context for later blocks and turns.
In multi-turn understanding, for example, later turns condition on earlier text answers by attending to their clean copies (Figure~\ref{fig:mask}d).
The clean copy is trimmed to the natural answer length, so padding appears only in the noisy answer, and it shares the position IDs of its noisy counterpart.

\paragraph{Visual answers.}
A visual answer has a similar structure: it starts with \texttt{<BOA>}\,\texttt{<BOI>}, followed by the grid of visual tokens with row separators, and ends with \texttt{<EOI>}\,\texttt{<EOA>}.
Its tokens are denoised simultaneously rather than block by block, and no later turn conditions on it, so it needs no clean copy.
A generation sample therefore takes the form $[\,C \mid \tilde V\,]$: the clean context $C$ followed by the noisy visual answer $\tilde V$.

\paragraph{Task-aware attention mask.}
The attention mask follows from the role of each token.
At a high level, a noisy answer is visible only to tokens of the same answer, visual answers attend bidirectionally, and text answers follow block diffusion.
More precisely, for each query--key pair in a packed sequence, the mask applies the four rules below.
Figure~\ref{fig:mask} illustrates these rules.
\begin{enumerate}
\item \textbf{Conditions are visible to their own and later turns; answers are invisible to conditions.} Any query may attend to the condition of its own turn and to clean keys from earlier turns; a clean query never attends to a noisy key.
\item \textbf{Text answers use turn-level block diffusion.} A noisy text query attends bidirectionally within its own block ($M_{\mathrm{BD}}$), reads the clean copy of earlier blocks of the same turn ($M_{\mathrm{OBC}}$), and cannot see later blocks. The clean copy is itself block-causal ($M_{\mathrm{BC}}$).
\item \textbf{Visual answers use spatial masked diffusion.} A noisy visual query attends to every noisy key of its own answer, to the condition of its own turn, to the clean copies of preceding answers such as the thinking text before an image, and to the clean context of earlier turns; there is no clean copy.
\item \textbf{Packed samples do not see each other.} Several samples are packed into one training sequence, and a query may attend to a key only if the two tokens belong to the same sample and neither is padding. The three rules above apply only to pairs that pass this check.
\end{enumerate}

\subsection{Mask Schedule and Training Objective}
\label{sec:train_obj}

With the above data format and the task-specific mask mechanism, all tasks can be optimized with the same masked-diffusion cross-entropy objective. In this section, we discuss the mask schedule and training objective of our Intern Lumina U2.

\paragraph{Mask Schedule for Text Modality}
Since the target text is generated through block diffusion at inference time, we design a block-wise mask schedule for training text generation to align this process.
Specifically, each target text is split into blocks of $B{=}32$ tokens, the block length used for all checkpoints in this report.
The mask ratio of each block is sampled independently: with probability $p_{\mathrm{full}}{=}0.03$, the block is fully masked; otherwise, we draw $u\sim\mathcal{U}(0,1)$ and mask a fraction $\rho=\cos(\tfrac{\pi}{2}u)$ of its maskable tokens, selected uniformly at random.
Special tokens are never masked, except \texttt{<EOA>} and \texttt{</think>}, which the model must learn to generate.

\paragraph{Mask Schedule for Visual Modality}

In contrast, all positions of the visual target share a single mask ratio: we randomly mask a fraction $\rho=\cos(\tfrac{\pi}{2}u)$ of its positions.
All $\Kcb$ codes at a masked position are supervised, and the loss of a masked position $j$ is the average of the $\Kcb$ per-codebook cross-entropy produced by the multi-codebook AR head:
\begin{equation}
  \ell_{\mathrm{vis}}(j) \;=\; \frac{1}{\Kcb}\sum_{k=1}^{\Kcb}
  -\log p\!\left(\code{k}^{\mathrm{gt}}\,\middle|\,\hbf,\, c^{\mathrm{gt}}_{<k}\right),
  \label{eq:visloss}
\end{equation}
where $\hbf$ and $\cbf^{\mathrm{gt}}$ are the hidden state and the ground-truth codes at position $j$, so that $\ell_{\mathrm{vis}}(j)$ equals $-\tfrac{1}{\Kcb}\log p(\cbf^{\mathrm{gt}}\mid\hbf)$ under Eq.~\eqref{eq:depthar}.
In other words, it is the negative log-likelihood of all $\Kcb$ codes, divided by $\Kcb$ so that a visual position carries the same weight as a text token.
In addition, to enable classifier-free guidance~\citep{Ho2022_CFG} at inference time, the condition of each text-to-image or editing sample is replaced by the \texttt{<uncondition>} token with probability $0.1$.

\paragraph{MoE load balancing.}
Because the backbone~\citep{llada2} routes each token to 8 of its 256 experts, maintaining a balanced load across experts is also essential during training, and we combine two mechanisms for this purpose.
The first is a differentiable load-balancing loss $\mathcal{L}_{\mathrm{balance}}$, which is included in the total loss below. The second follows the auxiliary-loss-free scheme of DeepSeek-V3~\citep{DeepSeekAI2024_V3}: each MoE layer keeps one bias value per expert, which is added to the router scores only when the experts are chosen. After each optimizer step, experts that received more tokens than the average have their bias decreased by $0.001$, and all other experts have it increased by $0.001$, so that overloaded experts are chosen less often in the following steps and underused experts more often.

\paragraph{Total loss.}
Text and visual losses share a single normalization. Let $\mathcal{T}$ be the set of masked text-answer tokens and $\mathcal{V}$ be the set of masked visual positions in the micro-batches that one data-parallel rank processes in a training step. Only these positions carry targets. $\mathcal{T}$ includes every masked \texttt{<EOA>} and \texttt{</think>}. Condition tokens, clean copies, the remaining special tokens, the boundary tokens of visual answers, and sequence padding carry no target. The total loss averages the per-token losses over $\mathcal{T}\cup\mathcal{V}$ and adds the load-balancing loss with weight $10^{-3}$:
\begin{equation}
  \mathcal{L} \;=\;
  \frac{\sum_{i\in\mathcal{T}}\ell_{\mathrm{txt}}(i) \;+\;
        \sum_{j\in\mathcal{V}}\ell_{\mathrm{vis}}(j)}
       {|\mathcal{T}| + |\mathcal{V}|}
  \;+\; 10^{-3}\,\mathcal{L}_{\mathrm{balance}},
  \label{eq:total}
\end{equation}
where $\ell_{\mathrm{txt}}(i)$ is the cross-entropy between the prediction at text position $i$ and its ground-truth token, and $\ell_{\mathrm{vis}}(j)$ is given by Eq.~\eqref{eq:visloss}. Each rank computes this loss over its own tokens, and the gradients are then averaged across data-parallel ranks.

\section{Data Construction}
\label{sec:data}

To train Intern Lumina U2, we construct a comprehensive and diverse training corpus and organize it in a consistent format.
For diversity, our training data span six categories: image understanding, video understanding, 3D understanding, text-to-image generation, image editing, and text-only data.
Each category is built from dedicated sources and filtered with category-specific quality checks.
For consistency, we convert all samples into a common dialogue format.
The converted samples are then mixed in stage-specific proportions for the training stages described in Section~\ref{sec:training}.

\subsection{Data Taxonomy and Unified Protocol}
\label{sec:data_protocol}

Although the training data differ in content, they share a common structure: each sample consists of conditions and targets.
In understanding data, the conditions include an image, a video, or a 3D asset, and the target is a text response; in generation data, the target is an image, generated either from a text prompt or by editing a source image according to an instruction.
To include both formats uniformly, we store every sample as a dialogue format labeled with a task type, such as image understanding, text-to-image generation, or multi-turn understanding.
The task type determines the ordering of conditions, noisy targets, and clean copies within a packed sequence, as described in Section~\ref{sec:seq_mask}.
As a result, a new data source can be incorporated simply by assigning it an existing task type, without modifying the sequence construction.

\begin{figure}[!ht]
  \centering
  \includegraphics[width=\linewidth]{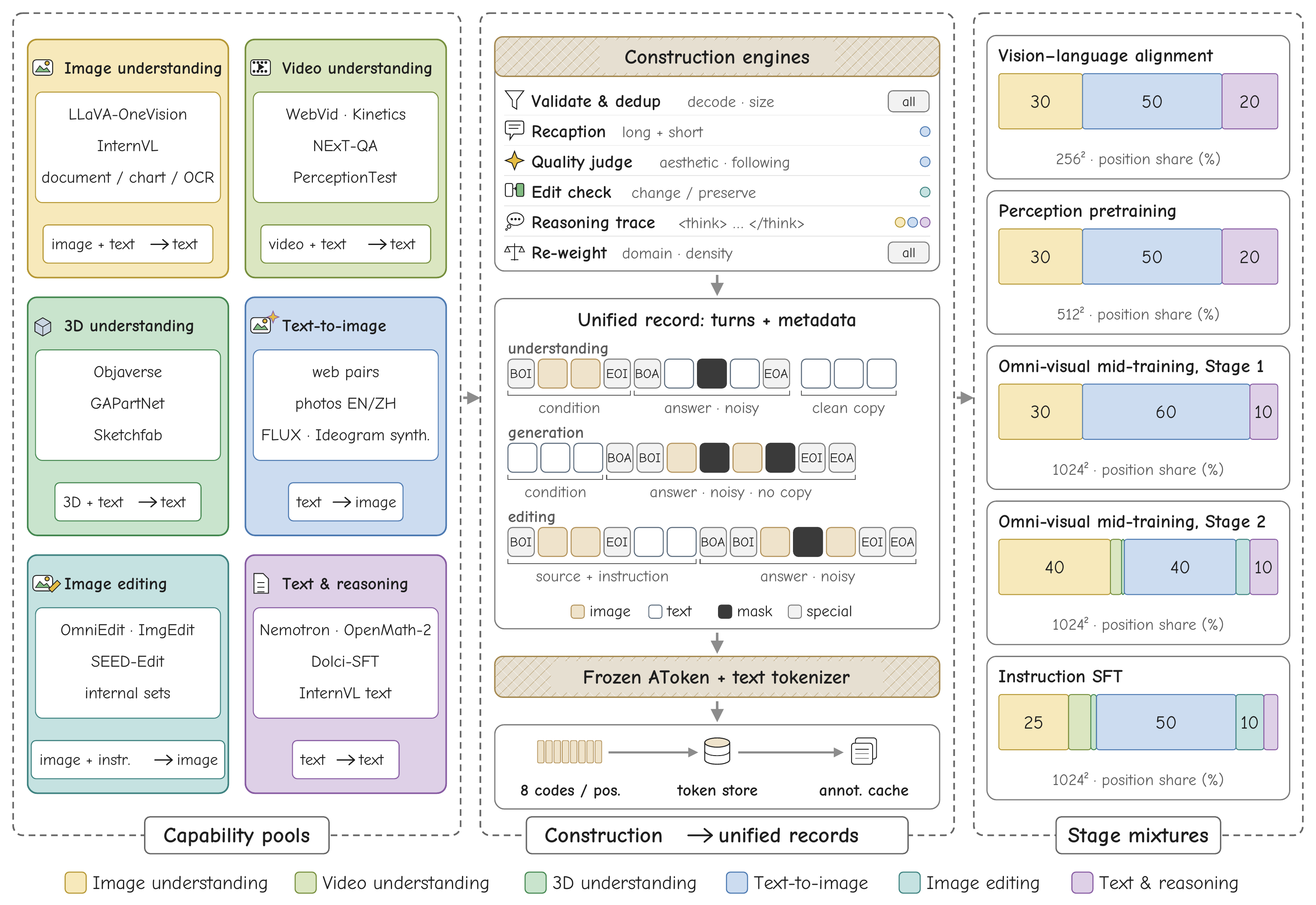}
  \caption{Data construction pipeline. Six data categories are built with category-specific rules, converted to one record format, encoded offline, and re-weighted at each training stage. Bars show backbone-position shares (Table~\ref{tab:data_versions}).}
  \label{fig:data_pipeline}
\end{figure}

The data mixture follows the five-stage curriculum described in
Section~\ref{sec:training}: Vision--Language Alignment, Perception
Pretraining, two stages of Omni-Visual Mid-Training, and Instruction SFT.
Editing, video, and 3D data are introduced in Omni-Visual Mid-Training
Stage~2. The final Instruction SFT stage uses a high-quality
instruction-tuning mixture curated from earlier data sources, covering
text-to-image tasks, image editing, omni-visual understanding across images,
videos, and 3D assets, and text-only question answering. Table~\ref{tab:data_versions}
summarizes the position-level shares across these stages, while the
corresponding stage configurations, sampling ratios, consumed tokens, and
training lineage are reported in Section~\ref{sec:training}.

\begin{table}[!htbp]
  \centering
  \caption{Data mixture at each training stage. Shares are fractions of estimated backbone positions per pass over the mixture; the corresponding resolution and optimization settings are given in Table~\ref{tab:stages}.}
  \label{tab:data_versions}
  \small
  \renewcommand{\arraystretch}{1.10}
  \setlength{\tabcolsep}{3pt}
  \begin{tabularx}{\linewidth}{@{}>{\centering\arraybackslash}p{0.40\linewidth}*{6}{>{\centering\arraybackslash}X}@{}}
    \toprule
    \multirow{2}{*}{\centering Stage} & \multicolumn{6}{c}{Position share (\%)} \\
    \cmidrule(lr){2-7}
    & Und. & T2I & Text & Edit & Video & 3D \\
    \midrule
    \mbox{Vision--Language alignment} & 30 & 50 & 20 & -- & -- & -- \\
    \mbox{Perception pretraining} & 30 & 50 & 20 & -- & -- & -- \\
    \mbox{Omni-visual mid-training, Stage 1} & 30 & 60 & 10 & -- & -- & -- \\
    \mbox{Omni-visual mid-training, Stage 2} & 40 & 40 & 10 & 5 & 4 & 1 \\
    \mbox{Instruction SFT} & 25 & 50 & 5 & 10 & 8 & 2 \\
    \bottomrule
  \end{tabularx}
\end{table}

\subsection{Text and Reasoning Data}
\label{sec:data_text}

Since our backbone is initialized from a language model, we include text-only data in every training stage to preserve its language and reasoning capabilities during multimodal training.
The text-only data contains three components: (i) public post-training corpora, namely the Nemotron post-training collections~\citep{Bercovich2025_LlamaNemotron} (covering chat, code, mathematics, and general instructions), OpenMathInstruct-2~\citep{Toshniwal2024_OpenMathInstruct2}, and Dolci-SFT~\citep{Olmo2025_Olmo3};
(ii) reasoning datasets with and without explicit chains of thought;
and (iii) the text-only portion of the private InternVL instruction data~\citep{Wang2025_InternVL35}.
We remove duplicate (user, assistant) pairs across all sources and reweight the sources by domain and by supervision density, i.e., the number of supervised tokens per sample.
The latter prevents short-response sources from dominating the mixture while contributing little to the loss.
Reasoning samples use the same format in the text, image understanding, and text-to-image data.
The reasoning trace is enclosed between \texttt{<think>} and \texttt{</think>} tags and followed by the final text response or image.

\subsection{Multimodal Understanding Data}
\label{sec:data_understanding}

Image understanding is the largest of the three understanding categories.
Its data are drawn mainly from the public instruction-tuning collection LLaVA-OneVision~\citep{Li2024_LLaVAOneVision} and the private instruction data of InternVL~\citep{Wang2025_InternVL35}, which includes image captioning, document understanding, and chart data.
To balance the mixture across skills, we organize these data by capability rather than by source dataset.
This yields three capability groups: perception (image captioning, general visual question answering, grounding, and counting), text-rich image understanding (OCR, documents, charts, and GUI or web screenshots), and knowledge-intensive reasoning (science questions and mathematics).

The video and 3D understanding data are limited in scale, but are constructed in the same manner as the image understanding data.
The video data comprise captioning and question-answering datasets built on public video corpora, together with temporal reasoning datasets covering event ordering, counting, and temporal localization tasks (see Appendix~\ref{app:data} for the full list of sources).
The 3D data consist of object-centric assets paired with captions and question-answering annotations, drawn from Objaverse~\citep{Deitke2023_Objaverse} (with Cap3D captions~\citep{Luo2023_Cap3D}), GAPartNet~\citep{Geng2023_GAPartNet}, and Sketchfab~\citep{Sketchfab2026}.

\subsection{Text-to-Image Data}
\label{sec:data_t2i}

Text-to-image generation requires training images that span a broad range of concepts and domains, contain rich details, exhibit high aesthetic quality, and are accurately described by their captions.
These requirements fall into three aspects: concept coverage, caption quality, and image quality.
We therefore organize the construction of the text-to-image corpus into three steps, each addressing one aspect: (i) source coverage and domain-specific expansion, which broaden the range of concepts and domains; (ii) caption reconstruction and prompt expansion, which improve caption quality; and (iii) filtering, deduplication, and balancing, which raise image quality and reduce redundancy.

\paragraph{Source coverage and domain-specific expansion.}
To cover a broad range of concepts, we draw on three types of sources: (i) web-crawled image--text pairs for the $256^2$ training stage; (ii) curated real-world photographs with captions in both English and Chinese; and (iii) synthetic images generated by existing text-to-image models, with the corresponding generation prompts used as captions.
Nevertheless, general-purpose sources alone cannot ensure comprehensive coverage: an analysis of generated samples revealed three weak areas, namely Chinese text rendering, scientific illustrations (including physics diagrams), and text-heavy graphic designs.
Rather than scaling up all sources, we construct a small dedicated dataset for each weak area and upweight it individually.

\paragraph{Caption reconstruction and prompt expansion.}
Because the prompt is the only conditioning signal in text-to-image generation, the quality of training prompts is critical.
We construct three types of prompts: original prompts, recaptioned prompts, and reasoning-augmented prompts.
Original prompts and alt-texts are retained only when they describe the image adequately.
In addition, InternVL3.5-30B-A3B~\citep{Wang2025_InternVL35} generates both a long-form and a short-form recaptioned prompt for images whose original captions are missing, mismatched, or too generic.
Both recaptioned prompts describe the subjects, attributes, relations, composition, style, and visible text of the image at different descriptive granularity.
During training, the prompt of each sample is randomly drawn from its original prompt and two recaptioned prompts.
Finally, for a subset of samples, we construct reasoning-augmented prompts by inserting a reasoning trace, enclosed in \texttt{<think>} tags, before the target image.
Within this trace, the prompt is expanded into an explicit description of the intended image.
These samples are used to train the thinking mode for image generation.

\paragraph{Filtering, deduplication, and balancing.}
This step consists of three stages: rule-based filtering and deduplication, model-based quality filtering, and data balancing.
In the first stage, every sample must pass validity checks that require a decodable image, a minimum side length, and a non-empty prompt with the length limit.
In the second stage, every sample is scored by a vision-language judge (Qwen3-VL-30B-A3B-Instruct~\citep{Bai2025_Qwen3VL}) for aesthetics, image quality, structural richness, and prompt following under the UniPercep protocol.
Filtering then applies three rules in order, discarding samples with illegible or missing required text, samples below the prompt-following threshold, and samples outside the top range of aesthetic scores; the thresholds and filtering statistics are given in Appendix~\ref{app:data}.
In the final stage, we balance the remaining data by adjusting the sampling multiplicity of each subset according to its content domain, language, and caption complexity.

\subsection{Image Editing Data}
\label{sec:data_editing}

Image editing is an image-to-image (I2I) task in which the model generates a target image conditioned on one or more source images with an editing instruction.
A valid editing sample must therefore satisfy two requirements: the target image reflects the requested change, while leaving unmentioned subjects, identities, layouts, and backgrounds unchanged.
Our editing data consist of two parts: public datasets, namely OmniEdit~\citep{Wei2024_OmniEdit}, ImgEdit~\citep{Ye2025_ImgEdit}, SEED-Data-Edit~\citep{Ge2024_SEEDDataEdit}, X2Edit~\citep{Ma2025_X2Edit}, and additional in-house synthetic data.
The resulting editing corpus covers ten edit types: addition, removal, replacement, adjustment, extraction, background change, style transfer, hybrid editing, action editing, and in-image text rendering.
Quality control follows directly from this definition, supplemented by a check on image quality.
Specifically, for each sample we verify that (i) the target differs visibly from the source; (ii) the observed change matches the instruction; (iii) regions not referenced by the instruction remain consistent between the source and target images; and (iv) the target image is of adequate visual quality.

\section{Training Strategy}
\label{sec:training}

The Intern Lumina U2 training pipeline contains five stages that progressively increase the length of the visual context, broaden the distribution of tasks, and expand the range of supported visual modalities.
At the start of training, the MoE backbone, text embedding table, and text prediction head are initialized from LLaDA2.0-mini~\citep{llada2}.
The visual embedding tables, visual input projection, and multi-codebook AR head are trained from scratch, while the AToken tokenizer remains frozen.
Each stage is initialized from the preceding checkpoint, and all stages use the unified objective of Section~\ref{sec:train_obj}.
The same training strategy is applied on two hardware platforms: the Huawei Ascend NPU run as the primary model, while the NVIDIA GPU run as a cross-platform recipe reference.

\subsection{Training Stages}
\label{sec:training_curriculum}

\paragraph{Stage 1: Vision--Language Alignment.}

Stage 1 aligns the random initialized visual modules (the visual embedding tables, visual input projection, and multi-codebook AR head) with the pretrained language backbone.
The model is trained at $256\times256$ resolution on a mixture of text-only, image captioning, and text-to-image data.
Image captioning and text-to-image data provide supervision for image-to-text and text-to-image generation, respectively, while text-only data preserve the language capabilities of the backbone.
Through these two complementary pathways, the model establishes an initial vision-language alignment and acquires preliminary image understanding and generation capabilities, which are further developed in the subsequent stages.

\paragraph{Stage 2: Perception Pretraining.}

Stage 2 increases the training resolution to $512\times512$ and broadens the
image understanding data to cover more demanding tasks, including OCR, grounding,
document understanding, and chart understanding.
The higher resolution provides the spatial detail required for fine-grained recognition and localization, particularly in text-rich documents and charts.
Consequently, the model acquires fine-grained perception capabilities, which are essential for the more complex visual tasks introduced in later stages.

\paragraph{Stage 3: Omni-Visual Mid-Training I.}
Stages 1 and 2 establish the basic vision--language alignment and the fundamental visual capabilities. The next two stages bring all visual modalities and capabilities from understanding to generation and editing into training. The first of them increases the training resolution to $1024\times1024$ without introducing new modalities or tasks.
Empirically, we find that expanding modality coverage simultaneously with the resolution increase leads to severe degradation of previously acquired capabilities.
We therefore use this stage to isolate the resolution transition: no new modalities or tasks are introduced while the model adapts its existing perception and generation capabilities to $1024\times1024$, allowing training to stabilize before modality coverage is expanded.

\paragraph{Stage 4: Omni-Visual Mid-Training II.}
Once training at $1024\times1024$ has stabilized, the second stage of omni-visual mid-training broadens the training recipe by incorporating data for image editing, video understanding, and 3D understanding.
The model learns these new tasks through the same visual interface and unified objective as in earlier stages.
Image and text data from the preceding stages remain in the mixture so that the model retains its existing capabilities.
The two stages of omni-visual mid-training share a combined token budget, which amounts to 500B tokens on the Ascend run and 200B tokens on the NVIDIA run, as reported in Table~\ref{tab:stages}.

\paragraph{Stage 5: Supervised Fine-Tuning for Instruction Following.}
This stage fine-tunes the model at $1024\times1024$ on a high-quality instruction-tuning dataset curated from the previous data sources and reformatted for instruction following.
The dataset covers text-to-image generation, image editing, omni-visual understanding (images, videos, and 3D assets), and text-only question answering.
Fine-tuning on these data aligns the capabilities acquired in earlier stages with instruction-following behavior, enabling the model to understand task instructions and produce the corresponding outputs.
The quality-filtered data further improve the overall performance of the model.
We also use a lower learning rate than in previous stages, so that the model gradually adapts to the instruction-following distribution while retaining the knowledge and generation quality it has already developed.

\begin{table*}[!htbp]
  \centering
  \begingroup
  \small
  \setlength{\tabcolsep}{4pt}
  \setlength{\arrayrulewidth}{0.3pt}
  \renewcommand{\arraystretch}{1.32}
  \newcommand{\tbc}{\textcolor{gray}{TBC}}
  \newcommand{\cfgsection}[1]{\multicolumn{6}{@{}l}{\cellcolor{black!6}\rule{0pt}{2.2ex}\textbf{#1}}\\}
  \begin{tabularx}{\linewidth}{@{}>{\raggedright\arraybackslash}p{0.28\linewidth}|*{4}{>{\centering\arraybackslash}X|}>{\centering\arraybackslash}X@{}}
    \hline
    \multirow{2}{*}{\textbf{Configuration}}
      & \multirow{2}{*}{\makecell{\textbf{Vision-}\\\textbf{Language}\\\textbf{alignment}}}
      & \multirow{2}{*}{\makecell{\textbf{Perception}\\\textbf{pretraining}}}
      & \multicolumn{2}{c|}{\makecell{\textbf{Omni-visual}\\\textbf{mid-training}}}
      & \multirow{2}{*}{\makecell{\textbf{Instruction}\\\textbf{SFT}}} \\
    & & & \textbf{Stage 1} & \textbf{Stage 2} & \\
    \hline
    \cfgsection{Training setup}
    Generation resolution & $256$ & $512$ & $1024$ & $1024$ & $1024$ \\
    Sequence length & 8,192 & 8,192 & 16,384 & 16,384 & 16,384 \\
    Text length limit & 2,048 & 2,048 & 2,048 & 2,048 & 2,048 \\
    Text block length & 32 & 32 & 32 & 32 & 32 \\
    Global batch size (Ascend) & 1,024 & 1,024 & 1,024 & 1,024 & 1,024 \\
    Global batch size (NVIDIA) & 192 & 192 & 192 & 192 & 192 \\
    Training tokens (Ascend) & $50$B & $200$B & \multicolumn{2}{c|}{$500$B (combined)} & $300$B \\
    Training tokens (NVIDIA) & $50$B & $200$B & \multicolumn{2}{c|}{$200$B (combined)} & $200$B \\
    \hline
    \cfgsection{Learning rates and regularization}
    Backbone peak LR & $2\!\times\!10^{-4}$ & $1\!\times\!10^{-4}$ & $1\!\times\!10^{-4}$ & $1\!\times\!10^{-4}$ & $1\!\times\!10^{-5}$ \\
    Text embedding/head LR & $1\!\times\!10^{-4}$ & $5\!\times\!10^{-5}$ & $5\!\times\!10^{-5}$ & $5\!\times\!10^{-5}$ & $5\!\times\!10^{-6}$ \\
    LR schedule & Cosine & Cosine & Cosine & Cosine & Constant \\
    Minimum backbone LR & $10^{-7}$ & $10^{-7}$ & $10^{-7}$ & $10^{-7}$ & $10^{-7}$ \\
    Warmup (\% of steps) & 3 & 3 & 3 & 3 & 1 \\
    Weight decay & 0.1 & 0.1 & 0.1 & 0.1 & 0.1 \\
    Norm/bias weight decay & 0 & 0 & 0 & 0 & 0 \\
    Gradient norm clip & 4.0 & 2.0 & 1.0 & 1.0 & 1.0 \\
    \hline
    \cfgsection{Optimizer}
    Matrix optimizer & Muon & Muon & Muon & Muon & Muon \\
    Fallback optimizer$^c$ & AdamW & AdamW & AdamW & AdamW & AdamW \\
    AdamW betas & $(0.9,\,0.999)$ & $(0.9,\,0.999)$ & $(0.9,\,0.999)$ & $(0.9,\,0.999)$ & $(0.9,\,0.999)$ \\
    AdamW epsilon & $10^{-8}$ & $10^{-8}$ & $10^{-8}$ & $10^{-8}$ & $10^{-8}$ \\
    Muon momentum & 0.95 & 0.95 & 0.95 & 0.95 & 0.95 \\
    Newton--Schulz steps & 5 & 5 & 5 & 5 & 5 \\
    Muon RMS scale & 0.2 & 0.2 & 0.2 & 0.2 & 0.2 \\
    Nesterov / per-head QKV & Yes / Yes & Yes / Yes & Yes / Yes & Yes / Yes & Yes / Yes \\
    \hline
  \end{tabularx}
  \par\vspace{3pt}
  \parbox{\linewidth}{\footnotesize $^c$ Muon is used for trainable two- and three-dimensional matrix parameters; AdamW handles embeddings, one-dimensional normalization and bias parameters, and the text output head and per-codebook output classifiers. Stage-wise learning rates and weight-decay settings are listed above.}
  \endgroup
  \caption{Training configurations for the primary Ascend run and the NVIDIA reference line.}
  \label{tab:stages}
\end{table*}

\subsection{Training Configurations}
\label{sec:training_configs}

Table~\ref{tab:stages} summarizes the stage-wise training configurations of the Ascend and NVIDIA, including the generation resolution, sequence length limits, batch sizes, token budgets, learning-rate and regularization schedules.
The proportions of the data mixture for each stage are provided in Table~\ref{tab:data_versions}.
On the Ascend platform, we use 1,024 Ascend 910B NPUs in all stages; on the NVIDIA platform, we use 192 GPUs in all stages.
On both platforms, the global batch size equals the number of devices, and Stages 1 and 2 process 50B and 200B tokens, respectively.
In the subsequent stages, the token budgets on the Ascend platform are 500B for omni-visual mid-training (Stages 3 and 4 combined) and 300B for SFT, compared with 200B for each on the NVIDIA platform.

\subsection{Optimization and Parallel Training}
\label{sec:training_system}

\paragraph{Optimizer.}
We optimize the trainable 2D and 3D weight matrices, including the attention, feed-forward, expert, and projection weights, with Muon~\citep{Jordan2024_Muon}, which has been shown to match AdamW with substantially fewer training tokens~\citep{Liu2025_MuonScalable} and to remain stable on trillion-parameter MoE models~\citep{Moonshot2025_KimiK2,KimiTeam2026_KimiK3}.
For the query, key, and value projections, we use the per-head variant of Kimi K3~\citep{KimiTeam2026_KimiK3}, which orthogonalizes the update of each attention head separately.
All other trainable parameters, namely the embedding layers, one-dimensional parameters such as norms and biases, and the text output head and the per-codebook output classifiers, are optimized with AdamW~\citep{Loshchilov2019_AdamW} using fixed betas $(0.9,0.999)$ and $\epsilon=10^{-8}$.
Stage-specific learning rates, weight decay, warm-up, gradient clipping, and Muon hyperparameters are reported in Table~\ref{tab:stages}.

\paragraph{Sequence packing and distributed training.}
We apply dynamic sequence packing in all training stages.
The data collator concatenates training samples until a predefined packing length limit is reached.
This removes only the padding inserted between samples, while retaining the padding within a single sample.
The data collator also records the cumulative sequence lengths of the packed samples, which mark the sample boundaries to confine attention and loss computation to each sample.
Therefore, this packing strategy reduces sequence-level padding overhead
while retaining sample-internal padding, without altering the objective for any individual sample.
On both platforms, we use FSDP2 with full parameter sharding in a hybrid sharded data parallel (HSDP) layout.
Gradient checkpointing is enabled, while parameter and activation offloading are disabled.
This configuration keeps long packed sequences within the per-device memory budget, while preserving identical model and loss definitions across hardware platforms.

\paragraph{Ascend and NVIDIA implementations.}
We train Intern Lumina U2 on both Huawei Ascend and NVIDIA platforms to evaluate the cross-platform effectiveness of our model.
The Ascend implementation is built on the CANN runtime, the VeOmni distributed training framework~\citep{Ma2025_VeOmni}, and HCCL for collective communication.
The attention and MoE layers are implemented with platform-native fused kernels; the Ascend-specific operator adaptations and memory constraints are detailed in Appendix~\ref{app:ascend}.
Both implementations follow the same five-stage schedule, unified training objective, and data proportions, while differing in hardware-specific settings, including device counts, parallelization meshes, and token budgets.
Because Ascend training uses a larger cluster and a larger token budget, we use the resulting checkpoint as the primary model for evaluation and deployment.
The model trained on NVIDIA GPUs serves as a cross-platform reference for the same training recipe.

\section{Inference}
\label{sec:inference}

\subsection{Inference Procedure}
\label{sec:inference_path}

\paragraph{Text decoding.}
We generate text with block diffusion~\citep{Arriola2025_BlockDiffusion,llada2}, which extends a fully decoded prefix from left to right while denoising the tokens within each appended block in parallel.
This design offers two practical advantages: (i) the generation process can be terminated at the \texttt{<EOA>} token, enabling variable-length outputs; and (ii) the key--value cache of the completed prefix can be reused when decoding the next block.

Given a prefix $x_{\leq p}$ of length $p$ and a block length $B$, decoding proceeds block by block, with $B$ mask tokens appended at a time.
Let $j$ denote the appended blocks, $s\in\{1,\ldots,S\}$ denote the denoising steps within a block, and $i$ denote the positions within a block.
At step $s$, the model samples a token at each masked position and assigns it a confidence $\sigma_i^{(j,s)}$, the predicted probability of the sampled token.
Following an unmask schedule $\{\nu_s\}_{s=1}^{S}$ with $\sum_{s=1}^{S}\nu_s=B$, the $\nu_s$ most confident positions are unmasked at step $s$, and the remaining positions keep masked:
\begin{equation}
  y^{(j,0)} = [x_{\leq p};\,\hat{x}^{<j};\,\mathrm{MASK}^{B}],
  \qquad
  \mathcal{A}_{j,s}=\operatorname{Top}_{\nu_s}\!\left(\{\sigma_i^{(j,s)}:i\in\mathcal{M}_{j,s}\}\right),
  \label{eq:text_block_decode}
\end{equation}
where $y^{(j,0)}$ is the input sequence at the start of block $j$, $\hat{x}^{<j}$ denotes the blocks decoded before block $j$, $\mathcal{M}_{j,s}$ is the set of positions of block $j$ that are still masked at the start of step $s$, with $\mathcal{M}_{j,1}$ containing all $B$ positions, $\mathcal{A}_{j,s}$ is the set of positions unmasked at step $s$, so that $\mathcal{M}_{j,s+1}=\mathcal{M}_{j,s}\setminus\mathcal{A}_{j,s}$, and $\operatorname{Top}_{\nu_s}$ returns the indices of the $\nu_s$ highest-confidence positions.
Unmasked tokens remain visible in subsequent steps, so the remaining masked tokens are predicted with bidirectional context within the block, while the blocks themselves are generated from left to right.
Once all positions in a block are generated, decoding stops if the block contains the \texttt{<EOA>} token, and all tokens after it are discarded; otherwise, another all-mask block is appended for generation.
Algorithm~\ref{alg:infer_text} summarizes the complete procedure, and Appendix~\ref{app:tokens} specifies the sequence format and termination conventions.

\begin{algorithm}
\small
\caption{Text inference (block diffusion)}
\label{alg:infer_text}
\begin{algorithmic}[1]
\Require prefix $x_{\leq p}$ (condition and \texttt{<BOA>}); generation length $G$; block length $B$; steps per block $S$; unmask counts $\nu_1,\ldots,\nu_S$ with $\sum_s \nu_s=B$; temperature $\tau$
\State $\hat x^{<0}\gets[\,]$
\For{$j=0,\dots,\lceil G/B\rceil-1$}
  \State $y^{(j,0)}\gets x_{\leq p}\,\Vert\,\hat x^{<j}$ (is\_clean${=}1$) $\Vert\,\texttt{<|mask|>}^{B}$ (is\_clean${=}0$, block $j$)
  \State build the attention mask from the token metadata
  \For{$s=1,\dots,S$}
    \State logits $\gets$ text head(backbone(current sequence)) at the masked positions $\mathcal{M}_{j,s}$
    \State $\hat x_i\gets\arg\max$(logits $+$ Gumbel noise $\cdot\,\tau$); $\sigma_i^{(j,s)}\gets$ softmax(logits)$[\hat x_i]$
    \State unmask the $\nu_s$ most confident positions of block $j$ ($\mathcal{A}_{j,s}$); keep the rest masked
  \EndFor
  \State $\hat x^{<j+1}\gets\hat x^{<j}\,\Vert$ block $j$; \textbf{stop} if block $j$ contains \texttt{<EOA>} (truncate there)
\EndFor
\State \Return the decoded blocks $\hat x$
\end{algorithmic}
\end{algorithm}

\paragraph{Visual decoding.}
Images are generated with full spatial diffusion: instead of proceeding block by block, all positions on the canvas are decoded jointly through iterative parallel refinement~\citep{Chang2022_MaskGIT}.
In each step, the backbone processes all masked positions in parallel using the global context of the canvas, and the multi-codebook AR head samples the $\Kcb=8$ codes at each position sequentially along the codebook index.
This preserves within-position code dependencies while allowing unresolved positions to be revised throughout generation.

Specifically, the canvas is initialized with all $N$ spatial positions masked.
Let $t\in\{1,\ldots,T\}$ denote the outer refinement steps, $i\in\{1,\ldots,N\}$ denote the spatial positions, and $k\in\{1,\ldots,\Kcb\}$ index the codebooks.
At step $t$, the backbone computes hidden states for the currently masked set $\mathcal{M}_t$ in parallel.
For each position $i\in\mathcal{M}_t$, conditioned on its head condition vector
$q_{t,i}=W_q\,\hbf_{t,i}+b_q$ and the previously sampled codes $\hat c_{t,i,<k}$, the multi-codebook AR head produces logits $z_{t,i,k}(a)$ over the codes of codebook $k$.
These logits define a categorical distribution over the candidate codes, from which the $\Kcb$ codes are sampled sequentially:
\begin{equation}
  \pi_{t,i,k}(a)
  =
  \operatorname{softmax}\!\left(z_{t,i,k}/\tau\right)_a,
  \qquad
  \hat c_{t,i,k}
  =
  \arg\max_a
  \left[
    z_{t,i,k}(a)+\tau\gamma_{t,i,k,a}
  \right],
  \qquad
  \sigma_{t,i}
  =
  \frac{1}{\Kcb}
  \sum_{k=1}^{\Kcb}
  \pi_{t,i,k}(\hat c_{t,i,k}),
  \label{eq:visual_decode_tuple}
\end{equation}
where $\tau>0$ is the sampling temperature, $\gamma_{t,i,k,a}$ is standard Gumbel noise, $\pi_{t,i,k}(a)$ denotes the probability assigned to code $a$ by the codebook-$k$ categorical distribution, and $\sigma_{t,i}$ is the mean probability of the sampled codes across the $\Kcb$ codebooks.
We utilize a cosine schedule, $n_t=\lfloor N\cos(\pi t/(2T))\rfloor$, to determine the number of positions kept masked after step $t$.
Under the baseline remasking rule, the $n_t$ positions with the lowest Gumbel-perturbed confidence scores remain masked, while the sampled codes at all other positions are committed and remain fixed.
This process is repeated until no masked positions remain, after which the AToken decoder reconstructs the image.
Algorithm~\ref{alg:infer_image} gives the complete sampling procedure.

\begin{algorithm}
\small
\caption{Image inference (spatial masked diffusion with multi-codebook AR decoding)}
\label{alg:infer_image}
\begin{algorithmic}[1]
\Require condition $C$ (prompt, plus the source image for editing) and unconditional condition $C_\varnothing$; grid of $N$ positions; refinement steps $T$; CFG scale $w$; temperature $\tau$; head condition projection $W_q,b_q$; position embeddings $s_k,u_k$ and code embeddings $E_k$; output classifiers $W_k^{\mathrm{out}}$
\State $V\gets$ all $N$ positions set to the mask entry (5000) in all $\Kcb$ codebooks (row delimiters and boundary tokens fixed)
\For{$t=1,\dots,T$}
  \State $\mathcal{M}_t\gets$ currently masked positions; $n_t\gets\lfloor N\cos(\tfrac{\pi}{2}\tfrac{t}{T})\rfloor$ (0 at $t{=}T$)
  \State $\hbf\gets(1+w)\,\text{backbone}([C\mid V])-w\,\text{backbone}([C_\varnothing\mid V])$ \Comment{guidance on hidden states}
  \State $q_{t,i}\gets W_q\hbf_{t,i}+b_q$ at positions $i\in\mathcal{M}_t$; prefix $\gets[q_{t,i}]$
  \For{$k=1,\dots,\Kcb$} \Comment{multi-codebook AR head, all positions in $\mathcal{M}_t$ in parallel}
    \State $g_{t,i,k}\gets$ head(prefix with $s_1,\dots,s_k$ added, adaLN from $q_{t,i}+u_k$); $z_{t,i,k}(a)\gets W^{\mathrm{out}}_k g_{t,i,k}$
    \State $\hat c_{t,i,k}\gets\arg\max_a(z_{t,i,k}(a)+\tau \gamma_{t,i,k,a})$; $\pi_{t,i,k}(\hat c_{t,i,k})\gets$ softmax$(z_{t,i,k}/\tau)[\hat c_{t,i,k}]$
    \State \textbf{if} $k<\Kcb$: prefix $\gets$ prefix $\Vert\,E_k(\hat c_{t,i,k})$
  \EndFor
  \State write $\hat c_{t,i,1:\Kcb}$ into $V$ at $i\in\mathcal{M}_t$; $\sigma_{t,i}\gets\frac{1}{\Kcb}\sum_k \pi_{t,i,k}(\hat c_{t,i,k})$
  \State re-mask $\min(n_t,|\mathcal{M}_t|)$ least confident positions of $\mathcal{M}_t$ (Gumbel-perturbed confidence with temperature $\tau$)
\EndFor
\State \Return AToken decoder($V$ without delimiters)
\end{algorithmic}
\end{algorithm}

\subsection{Inference Acceleration}
\label{sec:inference_accel}

Since image sampling is computationally expensive, we apply training-free acceleration at two levels of the sampler: (i) reducing the cost of each step by reusing cached intermediate states whose inputs remain stable, and (ii) adapting the remasking decisions when a position reaches a temporal plateau.
The following subsections describe these mechanisms and evaluate their quality--latency trade-off.

\subsubsection{Batching and Cache-Aware Reuse}
\label{sec:cache_reuse}

Recent masked-diffusion decoders reduce the repeated computation across refinement steps by reusing the intermediate
hidden representations or key-value caches.
Fast-dLLM~\citep{Wu2025_FastdLLM} and dKV-Cache~\citep{Ma2025_dKVCache} introduce approximate cache reuse for bidirectional diffusion decoders, and dLLM-Cache~\citep{Liu2025_dLLMCache} selectively refreshes cached response positions based on the V-verify score.
To reduce redundant computation, our inference implementation combines three complementary mechanisms for batch sharing and cache reuse.
First, the conditional and unconditional branches of classifier-free guidance (CFG) are evaluated together in one batch, which shares the backbone computation between the two branches.
Second, the multi-codebook AR head caches key-value states as it predicts the $\Kcb$ codes at each visual position, avoiding recomputation of the previously generated code prefix.
Third, the backbone reuses cached hidden states at positions with unchanged codes.
Together, these mechanisms reduce repeated computation in the CFG branches, the AR head’s code prefix, and backbone processing at positions where codes remain unchanged, while preserving the full $\Kcb$-step sequential prediction within each visual position.

Reusing cached backbone states requires a criterion for detecting stale cache entries.
For this purpose, we use a drift test adapted from the V-verify score of dLLM-Cache~\citep{Liu2025_dLLMCache}, to decide when the backbone cache must be refreshed.
At step $t$, let $\mathcal{S}_t$ denote the set of positions recomputed due to the change of input codes or the selection by the refresh schedule.
Let $h_{t,i,r}$ denote the conditional hidden-state component $r\in\{1,\ldots,d\}$ at position $i$, where $d$ is the hidden-state width. We measure the normalized drift between the recomputed and cached states:
\begin{equation}
  D_t = \frac{\operatorname{mean}_{i\in \mathcal{S}_t,\,r}
  |h^{\mathrm{new}}_{t,i,r}-h^{\mathrm{cache}}_{t,i,r}|}
  {\max\bigl(\operatorname{mean}_{i\in \mathcal{S}_t,\,r}
  |h^{\mathrm{cache}}_{t,i,r}|,\varepsilon_v\bigr)},
  \label{eq:vverify}
\end{equation}
where $\varepsilon_v>0$ ensures numerical stability.
If $D_t$ exceeds a threshold $\delta_v$, the entire backbone cache is refreshed at the next step; warm-up and fixed-interval refreshes remain in effect regardless of $D_t$.
Table~\ref{tab:infer_accel} reports the resulting number of backbone forward passes and the cache hit ratios.
Cache warm-up is included in the reported latency, so the initial forwards that build the cache are timed as part of each request.
Only the one-time kernel preheat used to initialize the kernels before timing is discarded.

\begin{table}[t]
  \centering
  \small
  \renewcommand{\arraystretch}{1.2}
  \caption{Inference acceleration under a matched $1024\times1024$ text-to-image
  protocol with the same checkpoint, 553 GenEval prompts, one image per prompt,
  CFG scale $4.0$, temperature $0.9$, and seed $42$. ``Steps'' is the realized
  number of outer diffusion steps, prompt-dependent for the plateau rows and
  fixed elsewhere. ``BB fwd'' counts conditional plus unconditional backbone
  forwards, with the cache-hit ratio in parentheses where a cache is active.
  Latency is wall-clock per image after a discarded kernel preheat and includes
  cache warm-up. The selected PHV setting is $\delta_p=0.18$, $f_p=0.005$,
  with the remaining cache and V-verify controls fixed across rows. GenEval uses one image per prompt and is therefore not
  comparable to the four-seed protocol of Table~\ref{tab:t2i}; no configuration
  scores below the production path.}
  \label{tab:infer_accel}
  \setlength{\tabcolsep}{6pt}
  \begin{tabular}{lrrrrr}
    \toprule
    Method & Steps & BB fwd & Lat.\,(s) & Speedup & GenEval \\
    \midrule
    \multicolumn{6}{l}{\emph{Cumulative system optimizations (fixed 96-step schedule)}} \\
    Production path & 96.0 & 192 & 159.9 & 1.00$\times$ & 86.6 \\
    \quad + CFG batching & 96.0 & 192 & 117.2 & 1.36$\times$ & 89.0 \\
    \quad + AR-head KV cache & 96.0 & 192 & 84.2 & 1.90$\times$ & 88.6 \\
    \quad + hidden-state cache & 96.0 & 192 (65\%) & 64.1 & 2.49$\times$ & 89.2 \\
    \quad + V-verify refresh & 96.0 & 192 (63\%) & 64.5 & 2.48$\times$ & 89.3 \\
    \quad + Halton selection (HVV) & 96.0 & 192 (64\%) & 66.5 & 2.40$\times$ & 88.2 \\
    \midrule
    \multicolumn{6}{l}{\emph{Reduced and adaptive step budgets}} \\
    HVV, fixed 60 steps & 60.0 & 120 (63\%) & 42.6 & 3.75$\times$ & 89.7 \\
    Plateau only (no cache, no Halton) & 56.9 & 114 & 46.5 & 3.44$\times$ & 87.2 \\
    Plateau + V-verify & 56.2 & 112 (35\%) & 34.8 & 4.59$\times$ & 87.2 \\
    \ourrow \textbf{PHV} & \textbf{60.4} & \textbf{121 (41\%)} & \textbf{35.9} & \textbf{4.45$\times$} & \textbf{89.3} \\
    \bottomrule
  \end{tabular}
\end{table}

\subsubsection{Plateau-Halton Sampling with V-verify (PHV)}
\label{sec:phv}
\label{sec:confidence_stability}

\begin{figure}[t]
  \centering
  \includegraphics[width=\linewidth]{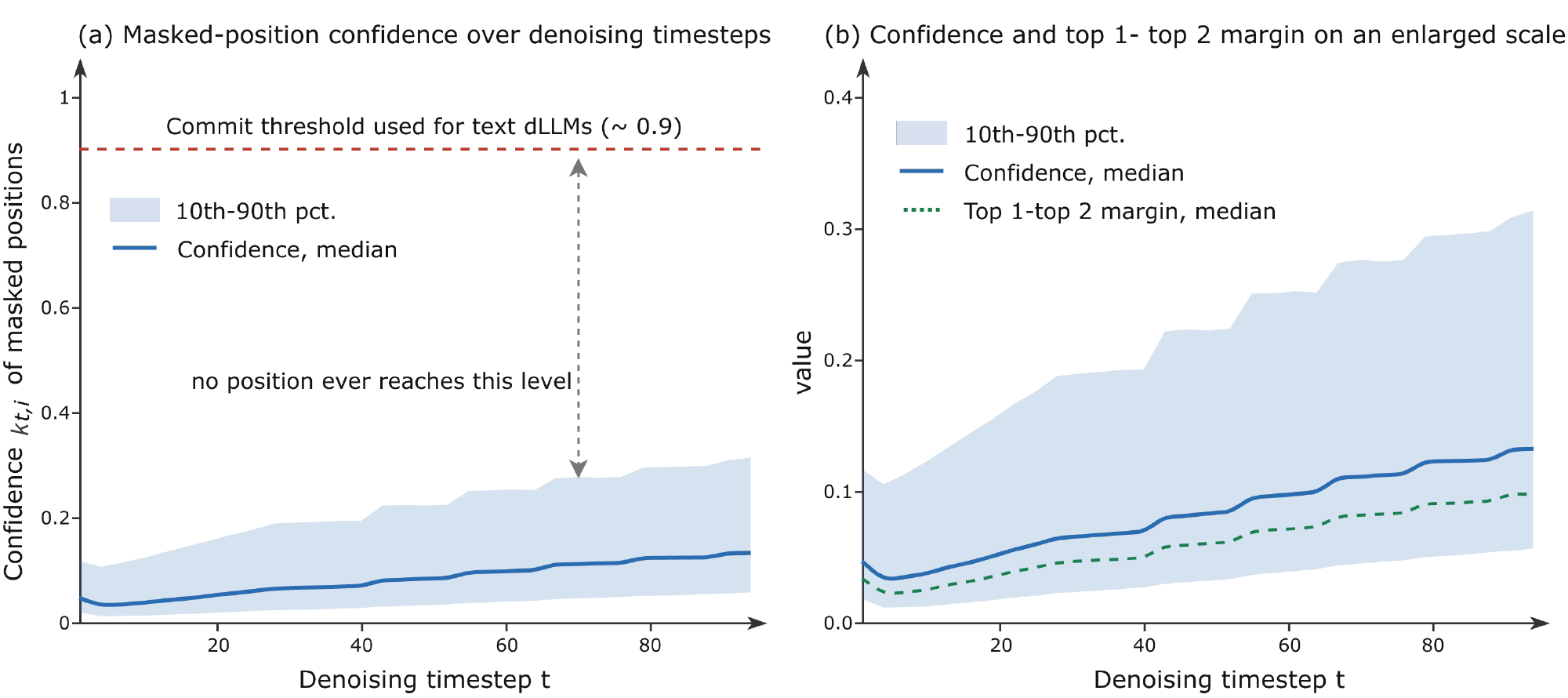}
  \caption{Confidence calibration for masked visual positions under the fixed
  $96$-step, $1024\times1024$ protocol. The visual confidence remains below the
  text-scale commit threshold, and its trajectory provides a more useful signal
  for adaptive sampling than its absolute level.}
  \label{fig:conf_calibration}
\end{figure}

\paragraph{Confidence calibration.}
As shown in Figure~\ref{fig:conf_calibration}, visual confidence exhibits both low magnitude and slow evolution: over $96$ refinement steps, its per-prompt median increases only from $0.046$ to $0.127$, and its aggregate 90th percentile stays below $0.32$.
A fixed confidence threshold is therefore ill-suited to this decoder.
We therefore propose Plateau-Halton sampling with V-verify (PHV), which uses confidence as a temporal signal for token commitment and pairs the sampler with the drift-triggered cache refresh of Section~\ref{sec:cache_reuse}.
For each masked position $i$ and codebook $k$, let
$\pi_{t,i,k}(a)$ denote the categorical probability assigned to code $a$ by the predictive logits at refinement step $t$.
We define the visual confidence at position $i$ as
\begin{equation}
  \kappa_{t,i}
  =
  \frac{1}{\Kcb}
  \sum_{k=1}^{\Kcb}
  \max_a \pi_{t,i,k}(a),
  \qquad
  \pi_{t,i,k}(a)
  =
  \operatorname{softmax}\!\left(z_{t,i,k}/\tau\right)_a,
  \label{eq:visual_confidence}
\end{equation}
where $\kappa_{t,i}$ is the average maximum code probability across the $\Kcb$ codebooks.
For remasking, PHV uses $\sigma_{t,i}$, the mean probability assigned to the sampled codes across the $\Kcb$ codebooks.

\paragraph{Plateau-Halton remasking.}
The sampler combines a temporal plateau test with a low-discrepancy Halton ordering~\citep{Besnier2025_Halton}, which evenly spreads the remaining uncertainty across the image.
For each position $i$, it maintains an exponential moving average (EMA) of the sampled confidence. With $\beta\in[0,1)$ and numerical stabilizer $\varepsilon_m>0$,
\begin{equation}
  m_{t,i}=\beta m_{t-1,i}+(1-\beta)\sigma_{t,i},
  \qquad
  \Delta_{t,i}=\frac{|m_{t,i}-m_{t-1,i}|}
  {\max(m_{t-1,i},\varepsilon_m)},
  \label{eq:plateau_ema}
\end{equation}
where $m_{t,i}$ is the EMA-smoothed confidence estimate at position $i$ after refinement step $t$, initialized to $\sigma_{t,i}$ at the first step in which position $i$ is sampled. A position becomes plateau-eligible once, after its first $n_{\min}$ samples, $\Delta_{t,i}$ has stayed below a threshold $\delta_p$ for $P$ consecutive steps and $m_{t,i}$ exceeds a floor $f_p$.
The floor $f_p$ prevents a flat but low-confidence trajectory from being treated as stable. We use $P=5$, $n_{\min}=4$, and $\beta=0.7$ throughout, and select $\delta_p$ and $f_p$ by the sweep in Section~\ref{sec:eval}.

At each step $t$, let $\mathcal{M}_t$ be the masked positions and let $\mathcal{U}_t\subseteq\mathcal{M}_t$ be those that are not plateau-eligible.
Plateau-eligible positions are committed immediately, and the cosine budget $n_t$ bounds how many of the remaining positions stay masked: the sampler retains the $r_t=\min(n_t,|\mathcal{U}_t|)$ positions $\mathcal{R}_t$ of $\mathcal{U}_t$ that come latest in a fixed Halton order of the grid and commits all other positions of $\mathcal{M}_t$, so that positions are committed in Halton order and spread evenly over the image.
The baseline sampler has no plateau test and keeps the $n_t$ positions with the lowest Gumbel-noised confidence masked, so the plateau-Halton sampler changes the eligibility criterion and the commit order while keeping the same budget and cache policy. Algorithm~\ref{alg:phv} gives the sampling procedure.

\begin{figure*}[t]
  \centering
  \includegraphics[width=\textwidth]{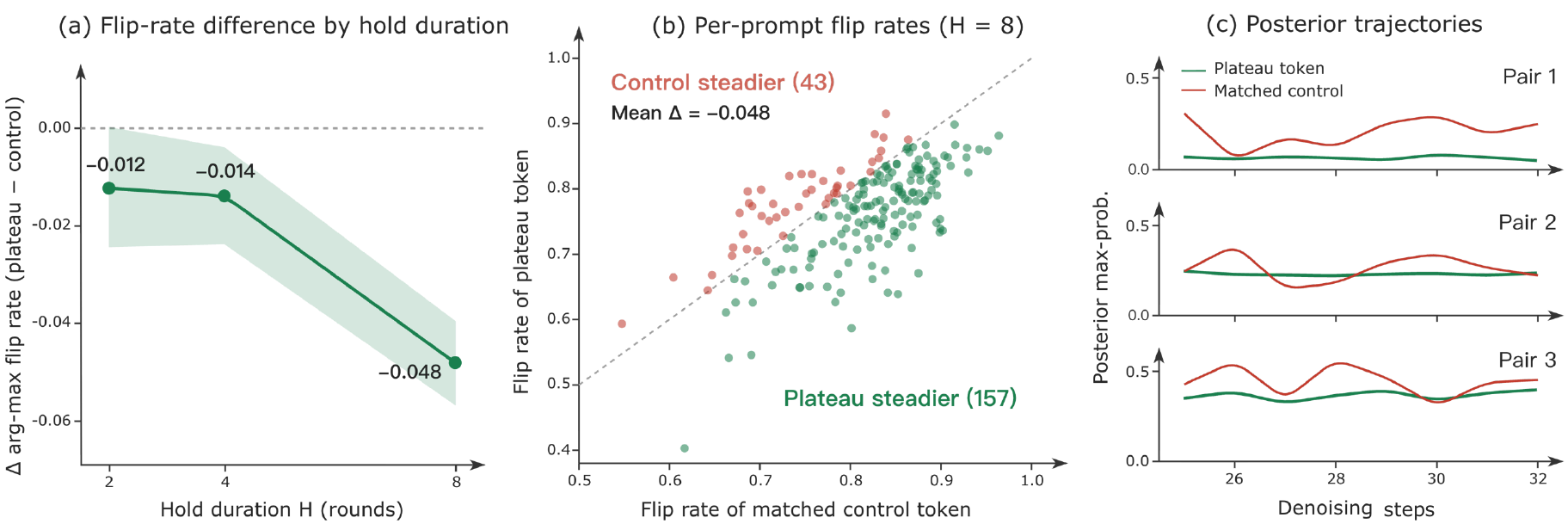}
  \caption{Offline temporal-stability test for PHV on $200$ prompts. Each
  plateau-eligible position is paired with a still-masked \emph{matched-control
  token} from the same image that does not satisfy the plateau criterion while
  exhibiting comparable confidence and EMA values. Both are held masked for
  $2$, $4$, or $8$ additional refinement steps. (a)~The difference in top-1 flip rate
  between the plateau token and its matched-control token grows with the hold
  duration.
  (b)~Per-prompt flip rates at $H=8$; $157$ of $200$ prompts fall below the diagonal.
  (c)~Posterior max-probability trajectories for three representative plateau--matched-control pairs.
  }
  \label{fig:hold_probe}
\end{figure*}

\paragraph{Temporal stability of PHV.}
PHV uses the plateau criterion online, while we assess its stability signal with
an offline holdout experiment. For a position that remains masked over
$H$ consecutive refinement transitions, we define the flip rate as the fraction
of codebook-wise top-1 predictions that change between adjacent steps:
\begin{equation}
  \operatorname{FlipRate}_{t,i}^{(H)}
  =
  \frac{1}{H\Kcb}
  \sum_{h=1}^{H}\sum_{k=1}^{\Kcb}
  \mathbf{1}\!\left[
    \arg\max_a\pi_{t+h,i,k}(a)
    \ne
    \arg\max_a\pi_{t+h-1,i,k}(a)
  \right].
  \label{eq:argmax_flip_rate}
\end{equation}
Here $\pi_{t+h,i,k}(a)$ is the predictive probability of candidate code $a$ for
position $i$ in codebook $k$ at refinement step $t+h$; $t$ is the initial
step of the measurement window, $H$ is the number of consecutive transitions,
and $\Kcb$ is the number of codebooks. The $\arg\max$ term selects the top-1
predicted code, and the indicator contributes one exactly when this code
changes between two adjacent steps. Lower values therefore indicate that the
predictive distribution changes less across refinement steps. To test whether
the plateau criterion identifies such positions, we pair each plateau-eligible
position with a still-masked position from the same image that does not satisfy
the plateau criterion while exhibiting comparable instantaneous confidence and
EMA values at the same step. We refer to the latter as a
\emph{matched-control token}.
Both tokens are then kept masked for $H\in\{2,4,8\}$ additional steps, so the
comparison controls for their confidence level while measuring subsequent
temporal change.
Across 200 prompts, the plateau tokens have lower flip rates than their
matched-control tokens. Figure~\ref{fig:hold_probe}(a) reports the mean
difference in flip rate, computed as plateau minus matched control: it changes
from $-0.012$ at $H=2$ to $-0.048$ at $H=8$ (95\% bootstrap CI
$[-0.057,-0.040]$ at $H=8$).
Figure~\ref{fig:hold_probe}(b) gives the per-prompt comparison at $H=8$:
the plateau token is steadier than its matched-control token for $157$ of the
$200$ prompts. Figure~\ref{fig:hold_probe}(c) provides qualitative examples
of the corresponding maximum-probability trajectories for three representative
pairs. Together, these results support an aggregate stability pattern rather than a
deterministic indicator of PHV convergence.

\begin{algorithm}[H]
\caption{PHV sampling for one visual answer}
\label{alg:phv}
\begin{algorithmic}[1]
\Require refinement steps $T$; temperature $\tau$; cosine budget $n_t$; plateau controls $(\delta_p,f_p,P,n_{\min},\beta)$; stabilizer $\varepsilon_m$
\State Mask all visual positions; initialize the plateau counters
\For{$t=1,\ldots,T$}
  \State $\mathcal{M}_t\gets$ still-masked positions; \textbf{if} $\mathcal{M}_t=\emptyset$ \textbf{then break}
  \State Obtain backbone states using the cache policy in Sec.~\ref{sec:cache_reuse}
  \State Sample a visual token ($\Kcb$ codes) for every $i\in\mathcal{M}_t$ with the multi-codebook AR head
  \State Update $m_{t,i}$, $\Delta_{t,i}$ and plateau counters for $i\in\mathcal{M}_t$
  \State $\mathcal{U}_t\gets\{i\in\mathcal{M}_t: i\ \text{not plateau-eligible}\}$
  \State $r_t\gets\min(n_t,\,|\mathcal{U}_t|)$ \Comment{cosine budget is a ceiling on what stays masked}
  \State $\mathcal{R}_t\gets$ the $r_t$ positions of $\mathcal{U}_t$ latest in the Halton order
         (PHV) or with the lowest Gumbel-noised confidence (baseline)
  \State Keep $\mathcal{R}_t$ masked; \textbf{commit} the sampled visual tokens at $\mathcal{M}_t\setminus\mathcal{R}_t$
\EndFor
\State Decode the completed grid of visual tokens with AToken
\end{algorithmic}
\end{algorithm}

\paragraph{Efficiency Protocol and Results.}
\label{sec:eff_protocol}

We evaluate the contribution of the cache policy, V-verify, and the plateau-Halton sampler under the same protocol specified in Table~\ref{tab:infer_accel}. The selected PHV configuration reduces latency from $159.9$ to $35.9$ seconds per image ($4.45\times$), while reducing the mean number of refinement steps from $96$ to $60.4$ and improving GenEval from $86.6$ to $89.3$. The gain over the plateau sampler with V-verify but without Halton ordering, which runs at a comparable latency, comes from the Halton reveal order: at the same step budget it improves which positions commit, raising GenEval by over two points. The sweep over the sampling parameters is reported in Section~\ref{sec:eval}.

\section{Evaluation}
\label{sec:eval}

We evaluate Intern Lumina U2 on image understanding and reasoning, video and 3D understanding, text-to-image generation, scientific-diagram generation, and image editing.
We report results for two models with the same architecture, tokenizer, and five-stage training schedule: the primary model, trained on 1,024 Huawei Ascend 910B NPUs (Appendix~\ref{app:ascend}), and a reference model trained on NVIDIA GPUs.
We then present ablations of the visual input interface, the multi-codebook AR head, and the inference-sampling configuration.

\paragraph{Setup.}
All benchmarks use the official evaluation sets and scoring protocols. Understanding benchmarks use 1024-pixel inputs, with chain-of-thought prompting on reasoning-heavy benchmarks such as MMMU~\citep{Yue2024_MMMU}, MathVista~\citep{Lu2024_MathVista}, and MathVision~\citep{Wang2024_MathVision}; video benchmarks~\citep{Fu2024_VideoMME,Zhou2024_MLVU,Li2024_MVBench} use 64 frames selected by segment-based random sampling. Generation and editing benchmarks follow their official multi-sample protocols and judges, namely GenEval~\citep{Ghosh2023_GenEval}, DPG-Bench~\citep{Hu2024_DPG}, UniGenBench~\citep{Wang2025_UniGenBench}, TIIF-Bench~\citep{Wei2025_TIIF}, WISE~\citep{Niu2025_WISE}, OneIG-Bench~\citep{Chang2025_OneIG}, ImgEdit~\citep{Ye2025_ImgEdit}, and GEdit-Bench~\citep{Liu2025_Step1XEdit}. Baselines include understanding-only models (Qwen2.5-VL~\citep{Bai2025_Qwen25VL} and LLaDA-V~\citep{You2025_LLaDAV}) and unified models (Janus-Pro~\citep{Chen2025_JanusPro}, Show-o2~\citep{Xie2025_Showo2}, BAGEL~\citep{Deng2025_BAGEL}, InternVL-U~\citep{Tian2026_InternVLU}, OmniGen2~\citep{Wu2025_OmniGen2}, MMaDA~\citep{Yang2025_6PQB9975}, Lumina-DiMOO~\citep{Xin2025_26M7KSHA}, LLaDA-o~\citep{You2026_LLaDAo}, and LLaDA2.0-Uni~\citep{Bie2026_LLaDA2Uni}); their numbers are taken from the sources listed in each table caption.

\begin{table}[!htbp]
  \centering
  \renewcommand{\arraystretch}{1.05}
  \setlength{\aboverulesep}{0pt}
  \setlength{\belowrulesep}{0pt}
  \footnotesize
  \caption{Image understanding and reasoning, grouped into document and chart understanding, perception and hallucination, and knowledge and mathematical reasoning.
  Benchmarks: ChartQA~\citep{Masry2022_ChartQA}, CharXiv~\citep{Wang2024_CharXiv}, HallusionBench~\citep{Guan2024_HallusionBench}, MME~\citep{Fu2023_MME}, RealWorldQA~\citep{xAI2024_RealWorldQA}, MMMU~\citep{Yue2024_MMMU}, MMMU-Pro~\citep{Yue2025_MMMUPro}, MathVista~\citep{Lu2024_MathVista}, MathVision~\citep{Wang2024_MathVision}, and DynaMath~\citep{Zou2024_DynaMath}.
  Baseline numbers are taken from \citet{Bie2026_LLaDA2Uni}.
  \textbf{Bold} marks the best score in each row.}
  \label{tab:understanding}
  \setlength{\tabcolsep}{4pt}
  \resizebox{\linewidth}{!}{%
  \begin{tabular}{lcccccccOO}
    \toprule
    & \multicolumn{2}{c}{Understanding-only} & \multicolumn{5}{c}{Unified} & \multicolumn{2}{>{\columncolor{ours_bg}}c}{Intern Lumina U2} \\
    \cmidrule(lr){2-3}\cmidrule(lr){4-8}\cmidrule(lr){9-10}
    Benchmark & \makecell{Qwen2.5-\\VL} & LLaDA-V & BAGEL & \makecell{Intern\\VL-U} & \makecell{Lumina-\\DiMOO} & LLaDA-o & \makecell{LLaDA\\2.0-Uni} & Ascend & NVIDIA \\
    \#Params & 7B & 8B & 7B+7B & 2B+1.7B & 8B & 8B & 16B-A1.4B & \multicolumn{2}{>{\columncolor{ours_bg}}c}{16.9B-A1.4B} \\
    \midrule
    \multicolumn{8}{l}{\emph{Documents and charts}} & \cellcolor{ours_bg} & \cellcolor{ours_bg} \\
    ChartQA & 84.1 & 78.3 & 74.3 & 76.6 & 8.3 & \textbf{87.9} & 80.1 & 86.2 & 81.2 \\
    CharXiv (DQ) & 73.9 & 47.0 & 70.6 & 53.3 & 27.8 & 69.8 & 68.4 & \textbf{85.9} & 84.0 \\
    \midrule
    \multicolumn{8}{l}{\emph{Perception and hallucination}} & \cellcolor{ours_bg} & \cellcolor{ours_bg} \\
    HallusionBench & 51.9 & 39.2 & 52.5 & 44.8 & 32.9 & 47.4 & 50.2 & \textbf{64.6} & 56.5 \\
    MME-C & 624 & 491 & \textbf{667} & 279 & 352 & 527 & 587 & \textbf{667} & 606 \\
    RealWorldQA & 68.5 & 63.2 & \textbf{73.9} & 56.4 & 52.4 & 60.8 & 66.7 & 62.9 & 59.9 \\
    \midrule
    \multicolumn{8}{l}{\emph{Knowledge and mathematical reasoning}} & \cellcolor{ours_bg} & \cellcolor{ours_bg} \\
    MMMU & 51.3 & 48.6 & 55.3 & 54.7 & \textbf{58.6} & 44.9 & 50.1 & 52.2 & 55.5 \\
    MMMU-Pro & 38.3 & 35.2 & 37.1 & 20.8 & 20.6 & 28.3 & 34.0 & \textbf{38.5} & 37.4 \\
    MathVista & 68.2 & 59.7 & \textbf{73.1} & 55.8 & 10.3 & 66.1 & 68.1 & 68.0 & 70.4 \\
    MathVision & 22.4 & 21.2 & 24.1 & 22.1 & 13.1 & 15.7 & 26.7 & \textbf{36.5} & 30.6 \\
    DynaMath & -- & -- & -- & 46.97 & -- & 40.58 & 46.89 & \textbf{63.6} & 55.8 \\
    \bottomrule
  \end{tabular}}
\end{table}

\paragraph{Image understanding and reasoning.}
Table~\ref{tab:understanding} reports the quantitative results on image understanding and reasoning benchmarks.
Despite having only 1.4B activated parameters, Intern Lumina U2 achieves the best score on six of the ten benchmarks (tied with BAGEL on MME-C), with the clearest leads on CharXiv, HallusionBench, MathVision, and DynaMath.
The comparison with LLaDA2.0-Uni is particularly informative because the two models share the same language backbone, whereas LLaDA2.0-Uni uses a semantic-only visual interface.
Intern Lumina U2 outperforms it on eight of the ten benchmarks, with the largest gains on benchmarks requiring fine-grained visual detail.
This pattern is consistent with the enhanced representational capacity of multi-codebook visual tokens.

\begin{table}[!htbp]
  \centering
  \renewcommand{\arraystretch}{1.05}
  \small
  \caption{Video and 3D understanding from the same checkpoint.
  Left: video benchmarks (Video-MME~\citep{Fu2024_VideoMME} without subtitles, MLVU~\citep{Zhou2024_MLVU} m-avg, MVBench~\citep{Li2024_MVBench}); baseline numbers are taken from \citet{Lin2023_VideoLLaVA,Zhang2025_LLaVAMini,Cheng2024_VideoLLaMA2,You2025_LLaDAV,Fu2024_VideoMME}.
  Right: 3D~MM-Vet~\citep{Qi2024_ShapeLLM}, with baseline numbers for ShapeLLM and PointLLM~\citep{Xu2024_PointLLM} taken from \citet{Qi2024_ShapeLLM}.
  \textbf{Bold} marks the best score in each column.}
  \label{tab:video3d}
  \footnotesize
  \setlength{\tabcolsep}{3pt}
  \begin{minipage}[t]{0.52\linewidth}
    \centering
    \begin{tabular}{lccc}
      \toprule
      Model & Video-MME & MLVU & MVBench \\
      \midrule
      Video-LLaVA-7B & 39.9 & -- & 43.1 \\
      LLaVA-Mini & -- & 42.8 & 44.5 \\
      VideoLLaMA2-7B & 46.6 & -- & 54.6 \\
      LLaDA-V & 56.1 & 59.5 & -- \\
      \ourrow Ours (Ascend) & 52.6 & 62.1 & \textbf{61.8} \\
      \ourrow Ours (NVIDIA) & \textbf{56.7} & \textbf{64.9} & 57.2 \\
      \bottomrule
    \end{tabular}
  \end{minipage}\hfill
  \begin{minipage}[t]{0.46\linewidth}
    \centering
    \begin{tabular}{lcccccc}
      \toprule
      Model & Rec & Know & Gen & Spat & Emb & Total \\
      \midrule
      PointLLM-7B & 40.6 & 49.5 & 34.3 & 29.1 & 48.7 & 41.2 \\
      ShapeLLM-7B & \textbf{45.7} & 42.7 & 43.4 & 39.9 & \textbf{64.5} & 47.4 \\
      \ourrow Ours (Ascend) & 34.6 & \textbf{57.2} & \textbf{48.2} & \textbf{60.1} & 56.8 & \textbf{50.5} \\
      \ourrow Ours (NVIDIA) & 32.0 & 49.9 & 43.2 & 37.9 & 57.9 & 43.8 \\
      \bottomrule
    \end{tabular}
  \end{minipage}
\end{table}

\paragraph{Video and 3D understanding.}
Table~\ref{tab:video3d} reports the quantitative results on video and 3D understanding benchmarks.
Notably, unlike Intern Lumina U2, none of the other unified baselines supports video and 3D understanding alongside image generation within a single checkpoint.
For video understanding, Intern Lumina U2 achieves the best score on all three benchmarks, ahead of the diffusion-based LLaDA-V and the dedicated video models such as VideoLLaMA2.
On 3D~MM-Vet, the Ascend-trained primary model also outperforms the 3D-specialist models ShapeLLM and PointLLM.
These results show that a single set of visual interfaces can support image, video, and 3D understanding at a level competitive with specialist models.

\begin{table}[!htbp]
  \centering
  \footnotesize
  \renewcommand{\arraystretch}{1.05}
  \caption{Text-to-image generation.
  GenEval~\citep{Ghosh2023_GenEval}, DPG-Bench~\citep{Hu2024_DPG}, TIIF-Bench~\citep{Wei2025_TIIF}, and WISE~\citep{Niu2025_WISE} overall scores (GenEval on a 0--100 scale); higher is better.
  Baseline numbers are taken from \citet{Bie2026_LLaDA2Uni}, \citet{Tian2026_InternVLU}, and \citet{Xie2025_Showo2}; TIIF reports the overall score on long prompts.
  UniGenBench results are reported per dimension in Table~\ref{tab:unigen_detail}.
  \textbf{Bold} marks the best unified-model score in each column}
  \label{tab:t2i}
  \setlength{\tabcolsep}{5pt}
  \begin{tabular}{llcccc}
    \toprule
    Model & \#Params & GenEval & DPG-Bench & TIIF & WISE \\
    \midrule
    \multicolumn{6}{l}{\emph{Generation-only models}} \\
    FLUX.1 [Dev] & 12B & 66 & 83.84 & 71.8 & 0.50 \\
    Qwen-Image & 20B & 87 & 88.32 & 86.8 & 0.62 \\
    \midrule
    \multicolumn{6}{l}{\emph{Unified understanding-and-generation models}} \\
    Janus-Pro & 7B & 80 & 84.19 & 65.0 & 0.35 \\
    Show-o2 & 7B & 76 & 86.14 & -- & 0.39 \\
    BAGEL & 7B+7B & 82 & 85.07 & 71.7 & 0.52 \\
    OmniGen2 & 3B+4B & 80 & 83.57 & -- & -- \\
    InternVL-U & 2B+1.7B & 85 & 85.18 & 73.9 & 0.46 \\
    MMaDA & 8B & 63 & 69.97 & -- & -- \\
    Lumina-DiMOO & 8B & 88 & 86.04 & 68.8 & 0.40 \\
    LLaDA-o & 8B & 86 & 87.04 & -- & -- \\
    LLaDA2.0-Uni & 16B-A1.4B & \textbf{89} & 87.76 & -- & \textbf{0.68} \\
    \ourrow Ours (Ascend) & 16.9B-A1.4B & \textbf{89} & 87.79 & \textbf{87.1} & 0.56 \\
    \ourrow Ours (NVIDIA) & 16.9B-A1.4B & \textbf{89} & \textbf{87.99} & 84.8 & 0.59 \\
    \bottomrule
  \end{tabular}
\end{table}

\begin{table}[!htbp]
  \centering
  \small
  \renewcommand{\arraystretch}{1.05}
  \caption{UniGenBench~\citep{Wang2025_UniGenBench} (EN) results on the ten first-level dimensions; Overall is their mean.
  Baseline numbers are taken from \citet{Bie2026_LLaDA2Uni}; \textbf{bold} marks the best unified-model score in each column.
  Dimensions: Sty(le), Wor(ld knowledge), Att(ribute), Act(ion), Rel(ationship), Log(ical reasoning), Gra(mmar), Com(pound), Lay(out), Txt (text generation).}
  \label{tab:unigen_detail}
  \setlength{\tabcolsep}{4pt}
  \resizebox{\linewidth}{!}{%
  \begin{tabular}{lcccccccccccc}
    \toprule
    Model & Sty & Wor & Att & Act & Rel & Log & Gra & Com & Lay & Txt & Overall \\
    \midrule
    \multicolumn{12}{l}{\emph{Generation-only models}} \\
    FLUX.1 [Dev] & 83.90 & 88.92 & 67.84 & 62.17 & 67.26 & 30.91 & 60.96 & 47.04 & 71.83 & 32.18 & 61.30 \\
    Qwen-Image   & 94.70 & 94.15 & 87.93 & 82.60 & 80.08 & 51.59 & 60.96 & 72.94 & 86.57 & 72.13 & 78.36 \\
    \midrule
    \multicolumn{12}{l}{\emph{Unified understanding-and-generation models}} \\
    Janus-Pro     & 90.80 & 86.71 & 67.74 & 64.26 & 68.40 & 37.05 & 64.44 & 62.11 & 72.01 & 2.59 & 61.61 \\
    BAGEL         & 90.20 & 85.60 & 67.74 & 61.98 & 70.69 & 30.23 & 66.44 & 58.12 & 76.49 & 7.76 & 61.53 \\
    OmniGen2      & 91.90 & 86.39 & 72.12 & 62.83 & 68.27 & 32.50 & 59.89 & 56.31 & 71.64 & 29.02 & 63.09 \\
    MMaDA         & 82.40 & 56.65 & 48.39 & 37.83 & 50.25 & 17.95 & 55.75 & 32.35 & 30.22 & 1.15 & 41.35 \\
    Lumina-DiMOO  & 89.70 & 90.03 & 81.62 & 71.12 & 78.43 & 45.45 & 70.45 & 73.32 & 82.84 & 25.57 & 71.12 \\
    LLaDA2.0-Uni  & 95.30 & 93.67 & 91.77 & 85.65 & 86.42 & 63.99 & 72.19 & 85.82 & 90.30 & \textbf{31.23} & 79.63 \\
    \ourrow Ours (Ascend) & \textbf{99.40} & \textbf{98.73} & \textbf{91.99} & \textbf{91.92} & \textbf{93.53} & \textbf{84.17} & \textbf{91.18} & \textbf{92.65} & \textbf{97.39} & 20.11 & \textbf{86.11} \\
    \ourrow Ours (NVIDIA) & 97.10 & 96.99 & 88.88 & 88.27 & 89.34 & 71.72 & 81.53 & 89.56 & 93.07 & 11.56 & 80.80 \\
    \bottomrule
  \end{tabular}}
\end{table}

\begin{table}[!htbp]
  \centering
  \small
  \renewcommand{\arraystretch}{1.05}
  \caption{TIIF-Bench~\citep{Wei2025_TIIF} results by instruction category for short and long prompts.
  Basic and Advanced are the average of their sub-categories; Real World is the Designer real-world score.
  Baseline numbers are taken from \citet{Tian2026_InternVLU}; \textbf{bold} marks the best unified-model score in each column.}
  \label{tab:tiif_detail}
  \setlength{\tabcolsep}{5pt}
  \begin{tabular}{lcccccccc}
    \toprule
    \multirow{2}{*}{Model} & \multicolumn{4}{c}{Short prompts} & \multicolumn{4}{c}{Long prompts} \\
    \cmidrule(lr){2-5}\cmidrule(lr){6-9}
    & Basic & Adv. & Real & Overall & Basic & Adv. & Real & Overall \\
    \midrule
    \multicolumn{9}{l}{\emph{Generation-only models}} \\
    FLUX.1 [Dev] & 83.1 & 65.8 & 70.7 & 71.1 & 78.7 & 68.5 & 71.5 & 71.8 \\
    Qwen-Image   & 86.2 & 79.3 & 90.3 & 86.1 & 87.2 & 80.9 & 91.4 & 86.8 \\
    \midrule
    \multicolumn{9}{l}{\emph{Unified understanding-and-generation models}} \\
    Janus-Pro    & 79.3 & 59.7 & 65.8 & 65.5 & 78.3 & 58.8 & 60.3 & 65.0 \\
    BAGEL        & 81.8 & 70.2 & 68.3 & 71.5 & 80.1 & 72.2 & 67.9 & 71.7 \\
    Lumina-DiMOO & 84.9 & 72.8 & 75.0 & 74.7 & 78.0 & 68.5 & 72.0 & 68.8 \\
    InternVL-U   & 82.3 & 73.5 & 65.3 & 74.9 & 81.5 & 72.7 & 66.8 & 73.9 \\
    \ourrow Ours (Ascend) & \textbf{92.53} & \textbf{86.03} & \textbf{94.40} & \textbf{88.57} & \textbf{92.94} & \textbf{85.02} & \textbf{93.28} & \textbf{87.08} \\
    \ourrow Ours (NVIDIA) & 86.71 & 81.68 & 92.54 & 84.59 & 90.32 & 83.97 & 88.43 & 84.77 \\
    \bottomrule
  \end{tabular}
\end{table}

\begin{table}[!htbp]
  \centering
  \small
  \renewcommand{\arraystretch}{1.05}
  \caption{OneIG-Bench~\citep{Chang2025_OneIG} (EN). Baseline numbers are taken from \citet{Bie2026_LLaDA2Uni}; Overall is the mean of the five dimensions.
  \textbf{Bold} marks the best unified-model score in each column.}
  \label{tab:oneig}
  \setlength{\tabcolsep}{6pt}
  \begin{tabular}{lcccccc}
    \toprule
    Model & Alignment & Text & Reasoning & Style & Diversity & Overall \\
    \midrule
    \multicolumn{7}{l}{\emph{Generation-only models}} \\
    FLUX.1 [Dev] & 0.786 & 0.523 & 0.253 & 0.368 & 0.238 & 0.434 \\
    Qwen-Image & 0.882 & 0.891 & 0.306 & 0.418 & 0.197 & 0.539 \\
    \midrule
    \multicolumn{7}{l}{\emph{Unified understanding-and-generation models}} \\
    Janus-Pro & 0.553 & 0.001 & 0.139 & 0.276 & \textbf{0.365} & 0.267 \\
    BAGEL & 0.769 & 0.244 & 0.173 & 0.367 & 0.251 & 0.361 \\
    OmniGen2 & 0.804 & 0.680 & 0.271 & 0.377 & 0.242 & 0.475 \\
    InternVL-U & 0.820 & \textbf{0.740} & 0.270 & 0.400 & 0.250 & 0.496 \\
    Lumina-DiMOO & 0.820 & 0.550 & 0.280 & 0.400 & 0.230 & 0.456 \\
    LLaDA2.0-Uni & \textbf{0.882} & 0.661 & \textbf{0.323} & 0.400 & 0.259 & \textbf{0.505} \\
    \ourrow Ours (Ascend) & 0.879 & 0.701 & 0.306 & \textbf{0.435} & 0.150 & 0.494 \\
    \ourrow Ours (NVIDIA) & \textbf{0.882} & 0.565 & 0.252 & 0.413 & 0.130 & 0.448 \\
    \bottomrule
  \end{tabular}
\end{table}

\paragraph{Image generation and editing.}
Tables~\ref{tab:t2i}--\ref{tab:oneig} report the results on text-to-image generation benchmarks.
Among unified models, Intern Lumina U2 achieves the best scores on DPG-Bench, UniGenBench, and TIIF-Bench, and matches the best score on GenEval.
It also outperforms the generation-only model FLUX.1~\citep{BFL2024_FLUX} and approaches the performance of Qwen-Image~\citep{Wu2025_QwenImage}, which activates about $14\times$ more parameters per token than our Intern Lumina U2.
In UniGenBench, its advantage is most pronounced in the dimensions of logical reasoning and grammar.
On TIIF-Bench, its performance remains stable across short and long instructions.
Without reasoning-augmented prompting, Intern Lumina U2 ranks second among unified models on WISE.
In OneIG-Bench, its overall score is comparable to the best unified model; it achieves the highest style score, but exhibits lower sample diversity than the baselines.
Table~\ref{tab:edit} reports image editing. On ImgEdit and English GEdit-Bench, Intern Lumina U2 is close to the best unified models and ahead of BAGEL and OmniGen2.

\paragraph{Scientific-diagram generation.}
Table~\ref{tab:scientific_diagrams} reports the results on two benchmarks assessing the physical correctness of generated physics diagrams.
GenExam Physics~\citep{Wang2025_GenExam} scores each diagram against fine-grained scoring points, whereas VeriphyT2IBench~\citep{Zhang2026_Princigram} verifies each diagram with item-specific binary questions at both the local and global levels.
On GenExam, Intern Lumina U2 obtains the highest relaxed score among general-purpose unified models, substantially outperforming SenseNova-U1 and BAGEL.
On VeriphyT2IBench, it attains the best Local score and the second-best Global score.
The per-domain breakdown in Table~\ref{tab:veriphy_breakdown} shows that this advantage is not driven by a single domain: Intern Lumina U2 ranks first in Local E\&M and Global mechanics, and performs comparably to the strongest physics-specialized model in the remaining domains.
Table~\ref{tab:veriphy_difficulty} further shows that its largest margin occurs on hard E\&M prompts, where it scores 81.30.

\begin{table}[!htbp]
  \centering
  \footnotesize
  \renewcommand{\arraystretch}{1.05}
  \caption{Physics-faithful scientific-diagram generation on the physics subset of GenExam and on VeriphyT2IBench.
  GenExam reports aggregate strict and relaxed MLLM-judge scores; VeriphyT2IBench reports question-level accuracy on its local attribute bank and global whole-diagram bank.
  Baseline values and protocol definitions follow \citet{Zhang2026_Princigram};
  Princigram uses physics-specific structured supervision and prompting, whereas the other rows are general-purpose generators.
  \textbf{Bold} marks the best score in each column.}
  \label{tab:scientific_diagrams}
  \setlength{\tabcolsep}{7pt}
  \begin{tabular}{lcccc}
    \toprule
    \multirow{2}{*}{Model} & \multicolumn{2}{c}{GenExam Physics} & \multicolumn{2}{c}{VeriphyT2IBench} \\
    \cmidrule(lr){2-3}\cmidrule(lr){4-5}
    & Strict & Relaxed & Local & Global \\
    \midrule
    \multicolumn{5}{l}{\emph{Reference and generation-only models}} \\
    Seedream 4.0 & 3.5 & 49.0 & 64.80 & 65.07 \\
    FLUX.1 Kontext max & 0.0 & 25.6 & -- & -- \\
    Gemini 2.5 Flash Image & -- & -- & 69.98 & 66.14 \\
    GPT-Image-1 & -- & -- & 68.32 & 68.56 \\
    Qwen-Image & 0.0 & 26.3 & 51.79 & 60.20 \\
    HiDream-I1-Full & 0.0 & 17.7 & 46.48 & 57.96 \\
    FLUX.1 dev & 0.0 & 14.4 & 45.18 & 56.32 \\
    SD 3.5 Large & 0.0 & 13.2 & 51.65 & 59.58 \\
    \midrule
    \multicolumn{5}{l}{\emph{Unified understanding-and-generation models}} \\
    BAGEL & 0.0 & 13.8 & 46.38 & 62.15 \\
    Lumina-DiMOO & -- & -- & 37.01 & 55.95 \\
    Show-o2-7B & 0.0 & 11.9 & 42.88 & 58.44 \\
    BLIP3o-NEXT-GRPO-Text-3B & 0.0 & 10.5 & 37.65 & 56.35 \\
    SenseNova-U1 & 0.0 & 23.2 & 61.90 & 64.84 \\
    \midrule
    \multicolumn{5}{l}{\emph{Physics-specialized variants}} \\
    Princigram-DiMOO & \textbf{8.0} & 50.4 & 72.05 & 77.25 \\
    Princigram-BAGEL & 5.3 & \textbf{54.8} & 75.69 & \textbf{82.54} \\
    \midrule
    \ourrow Intern Lumina U2 & 2.7 & 44.8 & \textbf{76.58} & 81.22 \\
    \bottomrule
  \end{tabular}
\end{table}

\begin{table}[!htbp]
  \centering
  \scriptsize
  \renewcommand{\arraystretch}{1.05}
  \setlength{\aboverulesep}{0pt}
  \setlength{\belowrulesep}{0pt}
  \caption{VeriphyT2IBench domain-level breakdown for representative models. Each panel reports question-level accuracy (percent; higher is better) for the local attribute bank or the global whole-diagram bank. The overall rows are omitted because the corresponding aggregate scores are already reported in Table~\ref{tab:scientific_diagrams}.}
  \label{tab:veriphy_breakdown}
  \setlength{\tabcolsep}{2.8pt}
  \begin{tabular}{lrrrrrrO}
    \toprule
    \emph{Local scoring} & Seedream & Gemini & GPT-Image & SenseNova & Princigram-D & Princigram-B & \makebox[2.2cm][c]{Intern Lumina U2} \\
    \midrule
    E\&M & 65.55 & 67.75 & 69.58 & 53.46 & 76.00 & 78.59 & \textbf{83.30} \\
    Mechanics & 56.19 & 61.63 & 60.25 & 52.16 & 68.24 & \textbf{71.59} & 71.34 \\
    Optics & 77.81 & \textbf{82.33} & 79.54 & 80.96 & 75.64 & 78.92 & 79.23 \\
    Quantum mechanics & 69.67 & 73.27 & 71.35 & 68.93 & 73.41 & \textbf{77.59} & 76.32 \\
    Thermodynamics & 42.91 & 58.90 & 49.24 & 46.24 & 60.40 & \textbf{66.95} & 63.73 \\
    Acoustics & 62.42 & 68.44 & 66.88 & 65.26 & 68.39 & \textbf{73.01} & 72.17 \\
    \midrule
    \emph{Global scoring} & Seedream & Gemini & GPT-Image & SenseNova & Princigram-D & Princigram-B & \makebox[2.2cm][c]{Intern Lumina U2} \\
    \midrule
    E\&M & 59.94 & 61.13 & 64.69 & 59.06 & 76.18 & \textbf{80.56} & 79.38 \\
    Mechanics & 66.00 & 65.48 & 68.83 & 62.74 & 80.47 & 83.40 & \textbf{84.80} \\
    Optics & 75.83 & 77.14 & 77.38 & 79.17 & 79.64 & \textbf{83.93} & 79.76 \\
    Quantum mechanics & 58.99 & 60.75 & 61.76 & 60.63 & 71.07 & \textbf{83.77} & 79.50 \\
    Thermodynamics & 66.82 & 67.63 & 72.60 & 67.86 & 77.46 & \textbf{83.47} & 81.62 \\
    Acoustics & 66.38 & 69.45 & 68.96 & 66.13 & 76.81 & \textbf{81.23} & 80.98 \\
    \bottomrule
  \end{tabular}
\end{table}

\begin{table}[!htbp]
  \centering
  \scriptsize
  \renewcommand{\arraystretch}{1.05}
  \setlength{\aboverulesep}{0pt}
  \setlength{\belowrulesep}{0pt}
  \caption{VeriphyT2IBench breakdown by prompt difficulty. Scores are percentages and higher is better. The panels report the six physics domains at easy, medium, and hard difficulty for the same representative models as Table~\ref{tab:veriphy_breakdown}.}
  \label{tab:veriphy_difficulty}
  \setlength{\tabcolsep}{2.8pt}
  \begin{tabular}{lrrrrrrO}
    \toprule
    \emph{Easy prompts} & Seedream & Gemini & GPT-Image & SenseNova & Princigram-D & Princigram-B & \makebox[2.2cm][c]{Intern Lumina U2} \\
    \midrule
    E\&M & 75.32 & 80.56 & 80.41 & 75.45 & 82.11 & \textbf{84.74} & 83.78 \\
    Mechanics & 66.08 & 67.84 & 68.19 & 64.96 & 74.41 & \textbf{76.77} & 75.40 \\
    Optics & 80.81 & \textbf{83.21} & 81.79 & 82.98 & 78.05 & 79.01 & 81.37 \\
    Quantum mechanics & 70.97 & 74.13 & 75.41 & 73.39 & 75.58 & 81.34 & \textbf{81.47} \\
    Thermodynamics & 64.75 & 66.77 & 67.66 & 64.61 & 68.93 & \textbf{72.13} & 69.30 \\
    Acoustics & 64.88 & 69.86 & 69.19 & 69.72 & 67.84 & 71.13 & \textbf{71.94} \\
    \midrule
    \emph{Medium prompts} & Seedream & Gemini & GPT-Image & SenseNova & Princigram-D & Princigram-B & \makebox[2.2cm][c]{Intern Lumina U2} \\
    \midrule
    E\&M & 68.70 & 71.33 & 72.64 & 53.65 & 79.35 & 83.28 & \textbf{84.20} \\
    Mechanics & 58.97 & 61.79 & 62.82 & 54.07 & 72.28 & \textbf{75.93} & 75.47 \\
    Optics & 76.11 & 80.51 & 79.34 & \textbf{81.03} & 75.60 & 77.80 & 75.62 \\
    Quantum mechanics & 70.16 & 74.05 & 70.04 & 69.88 & 73.46 & \textbf{78.37} & 74.47 \\
    Thermodynamics & 53.32 & 67.12 & 58.44 & 53.18 & 64.79 & \textbf{73.57} & 69.59 \\
    Acoustics & 65.36 & 72.93 & 69.32 & 68.38 & 70.50 & 73.33 & \textbf{74.55} \\
    \midrule
    \emph{Hard prompts} & Seedream & Gemini & GPT-Image & SenseNova & Princigram-D & Princigram-B & \makebox[2.2cm][c]{Intern Lumina U2} \\
    \midrule
    E\&M & 53.89 & 54.34 & 57.29 & 38.23 & 69.09 & 69.97 & \textbf{81.30} \\
    Mechanics & 50.67 & 58.73 & 55.70 & 45.68 & 64.74 & 68.34 & \textbf{69.46} \\
    Optics & 76.36 & \textbf{82.32} & 77.48 & 78.94 & 74.17 & 80.44 & 80.39 \\
    Quantum mechanics & 64.42 & 67.52 & 65.79 & 60.99 & 70.53 & \textbf{75.28} & 74.49 \\
    Thermodynamics & 34.29 & 53.13 & 42.48 & 40.44 & 58.68 & \textbf{64.99} & 62.41 \\
    Acoustics & 58.43 & 63.33 & 62.90 & 57.96 & 69.85 & \textbf{77.39} & 73.15 \\
    \bottomrule
  \end{tabular}
\end{table}

\begin{table}[!htbp]
  \centering
  \renewcommand{\arraystretch}{1.05}
  \footnotesize
  \caption{Instruction-based image editing.
  ImgEdit~\citep{Ye2025_ImgEdit} overall score (1--5) and GEdit-Bench~\citep{Liu2025_Step1XEdit} overall score $G_O$ (0--10) for English and Chinese prompts.
  Baseline numbers are taken from \citet{Bie2026_LLaDA2Uni} and \citet{Tian2026_InternVLU}.
  \textbf{Bold} marks the best unified-model score in each column.}
  \label{tab:edit}
  \setlength{\tabcolsep}{5pt}
  \begin{tabular}{llccc}
    \toprule
    Model & \#Params & ImgEdit & GEdit-EN & GEdit-CN \\
    \midrule
    \multicolumn{5}{l}{\emph{Editing-specialist models}} \\
    Step1X-Edit & 12B & 3.06 & 6.97 & 6.86 \\
    Qwen-Image-Edit & 20B & 4.35 & 7.56 & 7.52 \\
    \midrule
    \multicolumn{5}{l}{\emph{Unified understanding-and-generation models}} \\
    BAGEL & 7B+7B & 3.20 & 6.52 & 6.50 \\
    OmniGen2 & 3B+4B & 3.44 & 6.41 & -- \\
    InternVL-U & 2B+1.7B & 3.67 & \textbf{6.66} & -- \\
    Lumina-DiMOO & 8B & 2.77 & 3.91 & -- \\
    LLaDA2.0-Uni & 16B-A1.4B & \textbf{3.92} & 6.61 & \textbf{6.66} \\
    \ourrow Intern Lumina U2 (Ascend) & 16.9B-A1.4B & 3.82 & 6.06 & 5.95 \\
    \ourrow Intern Lumina U2 (NVIDIA) & 16.9B-A1.4B & 3.82 & 6.54 & 6.01 \\
    \bottomrule
  \end{tabular}
\end{table}

\begin{figure}[!htbp]
  \centering
  \includegraphics[width=\linewidth]{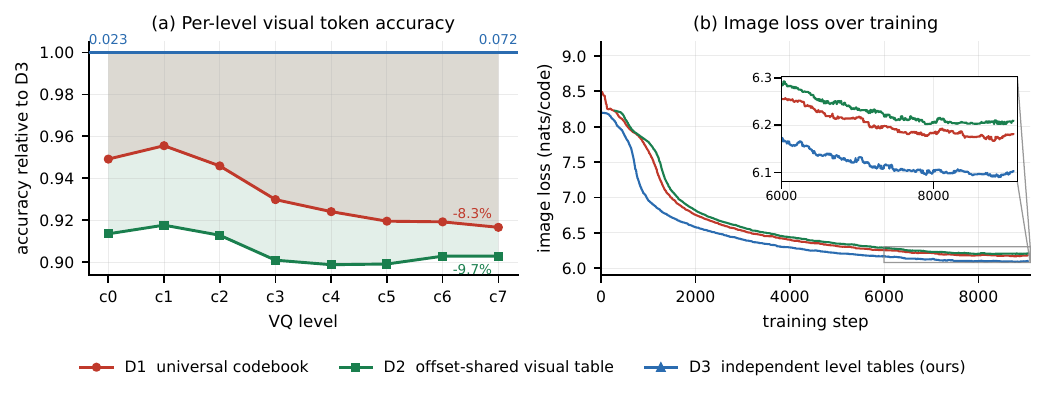}
  \caption{Input-interface ablation. \textbf{(a)} Final next-code top-1 accuracy for each of the eight codebooks, normalized by D3; the blue annotations give D3's absolute accuracy at the first and last codebooks. \textbf{(b)} Image loss over the two training epochs, shown as a rolling median with an inset for the converged tail. Higher is better in (a), whereas lower is better in (b).}
  \label{fig:input_ablation}
\end{figure}

\paragraph{Input-interface ablation.}
Table~\ref{tab:input_ablation} and Figure~\ref{fig:input_ablation} compare the three input designs illustrated in Figure~\ref{fig:input_designs}: (D1) a single embedding table shared by text tokens and visual codes; (D2) a single visual embedding table with codebook-specific index offsets; and (D3, ours) a separate embedding table for each codebook, followed by concatenation and linear projection.
Because each input design also determines the width of the corresponding output classifier, the comparison covers complete interface designs rather than input embeddings alone.
All other settings are kept identical.

D3 achieves the highest next-code accuracy, exceeding both D1 and D2 by more than ten standard errors, and the gap widens for later codebooks (Figure~\ref{fig:input_ablation}a).
The image-loss curves exhibit the same ordering after the first 2,000 training steps (Figure~\ref{fig:input_ablation}b).
D3 is also cost-effective: the large shared classifier of D1 requires $2.35\times$ the memory of D3 and leads to slower convergence during training, whereas D3 matches the memory of D2 with only $1.9\%$ more interface parameters.
D3 is therefore the preferred design in terms of both accuracy and efficiency.

\begin{table}[!htbp]
  \centering
  \renewcommand{\arraystretch}{1.05}
  \footnotesize
  \caption{Ablation of the visual input interface. Each design determines both the lookup-table organization and the corresponding classifier width. Accuracy is next-code top-1 on held-out supervised codes at the final checkpoint ($\approx5$M codes per arm; $\pm$ one standard error). Memory and step time are measured at sequence length $4608$ on identical hardware. \textbf{Bold} marks the best value in each column.}
  \label{tab:input_ablation}
  \setlength{\tabcolsep}{3pt}
  \begin{tabular}{@{}llccccc@{}}
    \toprule
    Design & Table organization & \makecell{Softmax\\width} & Params & \makecell{Peak memory\\(GB)} & \makecell{Step time\\(s)} & \makecell{Next-code\\accuracy} \\
    \midrule
    D1 universal & $1\times189{,}968$ & $189{,}968$ & \textbf{778.11M} & 43.79 & 7.1 & $0.0426\pm0.0002$ \\
    D2 offset-shared & $1\times32{,}784$ & $32{,}784$ & \textbf{778.11M} & 18.68 & \textbf{6.2} & $0.0415\pm0.0002$ \\
    \ourrow D3 independent (ours) & $8\times4{,}096$ & \textbf{4{,}096} & 793.02M & \textbf{18.62} & 6.6 & $\mathbf{0.0459\pm0.0002}$ \\
    \bottomrule
  \end{tabular}
\end{table}

\paragraph{Output-head ablation.}
Table~\ref{tab:head_ablation} compares the multi-codebook AR head with two alternative approaches under comparable training budgets at $512\times512$ resolution.
The first is the independent parallel prediction head.
The second is a Q-Former-style prediction head~\citep{Li2023_BLIP2}, which predicts all $\Kcb$ codes in a single pass from learned queries; like the parallel head, it treats the codes as conditionally independent.
Among the three designs of Section~\ref{sec:mcio}, flattening the codes into separate tokens is not included, because it multiplies the visual sequence length by $\Kcb$, which makes its training and inference cost impractical. In its place, we include a Q-Former-style head as a competitive alternative.

The multi-codebook AR head achieves the best results on all four benchmarks. The parallel head matches it on DPG-Bench but trails it by 14 points on GenEval, with the largest drops in counting, position, and attribute binding, which require a single precise configuration of objects and attributes.
This is consistent with the analysis in Section~\ref{sec:mcio}: since the $\Kcb$ codes are not conditionally independent given $\hbf$, ignoring their dependencies will degrade generation quality.
Despite its more expressive conditioning, the Q-Former head performs no better than the parallel head on GenEval and worse on the other three benchmarks.
This result indicates that the advantage of the multi-codebook AR head stems from modeling inter-code dependencies rather than from more expressive conditioning.

\begin{table}[!htbp]
  \centering
  \renewcommand{\arraystretch}{1.05}
  \small
  \caption{Ablation of the visual output head: three training runs at 512-pixel generation resolution with comparable token budgets (about 197B--220B tokens each).
  All scores are on a 0--100 scale; TIIF is the mean accuracy and UniGenBench the EN ten-dimension average.
  \textbf{Bold} marks the best score in each column.}
  \label{tab:head_ablation}
  \setlength{\tabcolsep}{5pt}
  \begin{tabular}{lccccc}
    \toprule
    Output head & \makecell{Codebook\\dependence} & GenEval & DPG-Bench & TIIF & UniGenBench \\
    \midrule
    Independent parallel & no & 68.94 & 83.22 & 70.1 & 59.6 \\
    Q-Former & no & 69.67 & 78.92 & 63.9 & 56.7 \\
    \ourrow Multi-codebook AR (ours) & yes & \textbf{82.78} & \textbf{83.27} & \textbf{73.2} & \textbf{62.1} \\
    \bottomrule
  \end{tabular}
\end{table}

\begin{figure}[h]
  \centering
  \includegraphics[width=0.92\linewidth]{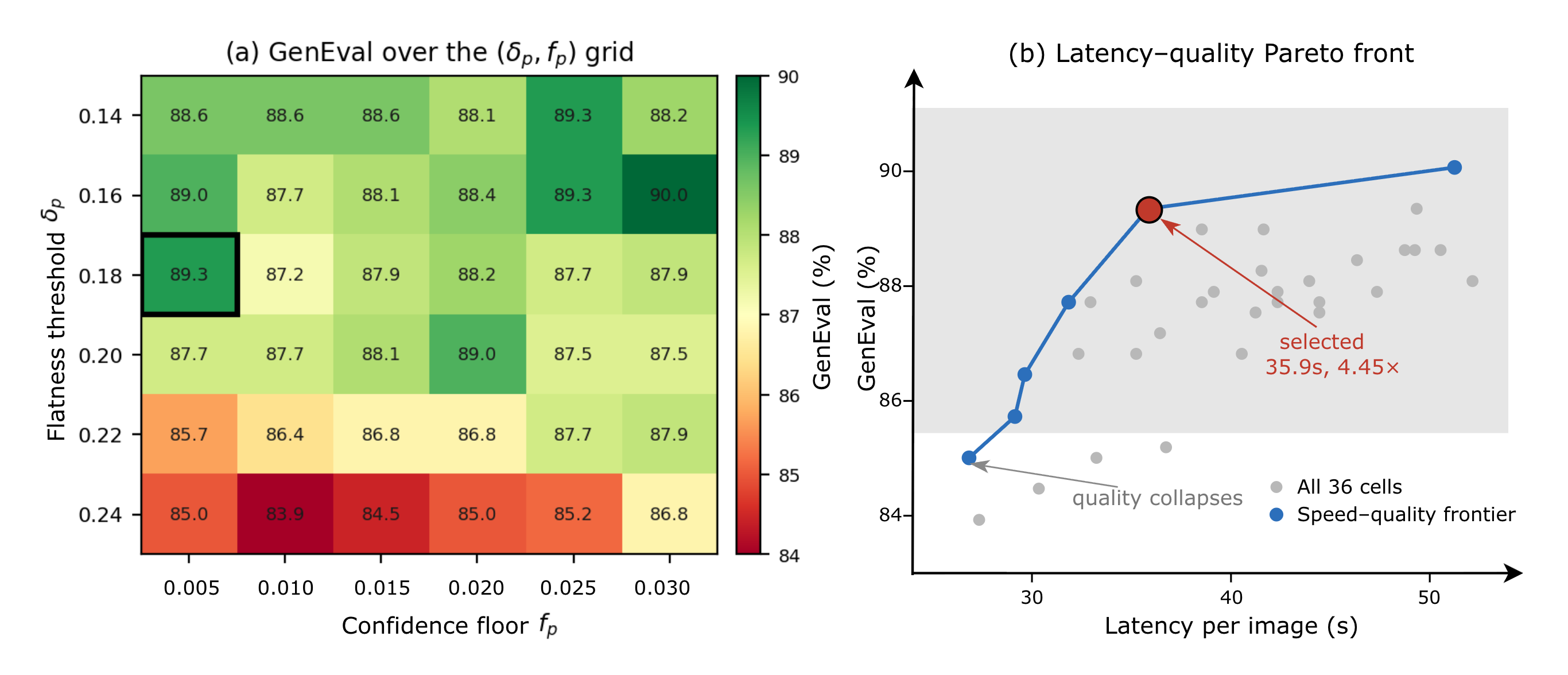}
  \setlength{\abovecaptionskip}{4pt}
  \caption{Inference-sampling ablation over the plateau change threshold $\delta_p$ and confidence floor $f_p$, $553$ prompts per cell.
  \textbf{(a)} GenEval over the $6\times6$ grid; the black box marks the selected quality-first setting $\delta_p{=}0.18$, $f_p{=}0.005$.
  \textbf{(b)} Speed--quality frontier over all $36$ cells, i.e.\ the configurations not improved on in both latency and quality by any other cell; the selected setting (red) reaches GenEval $89.3$ at $35.9$~s per image ($4.45\times$).}
  \label{fig:param_main}
\end{figure}

\paragraph{Inference-sampling ablation.}
We ablate the two plateau controls of PHV (Section~\ref{sec:phv}), the change threshold $\delta_p$ and the confidence floor $f_p$, on a $6\times6$ grid with $553$ prompts per cell (Figure~\ref{fig:param_main}); the remaining controls are fixed at their default values.
The two parameters act as a serial gate rather than as independent controls: $\delta_p$ determines which positions become eligible for early commitment, and $f_p$ only limits how many of the eligible positions commit.
As a result, the influence of $f_p$ on the number of refinement steps increases by $17.6\times$ from the strictest to the loosest $\delta_p$, since at small $\delta_p$ the flatness test rarely fires and $f_p$ has little effect.
The selected quality-first setting ($\delta_p{=}0.18$, $f_p{=}0.005$) lies on the speed--quality frontier, reaching GenEval $89.3$ at $35.9$~s per image.

\section{Conclusion}
\label{sec:conclusion}
We have presented Intern Lumina U2, to our knowledge the first unified understanding-and-generation diffusion language model that models multi-codebook visual states in a fully discrete space and extends visual understanding across images, videos, and 3D assets.
Its visual input interface preserves codebook identity by assigning each codebook an independent embedding table and merging the resulting embeddings into a single backbone token through concatenation and projection.
Its visual output interface uses a multi-codebook AR head to predict the codes at each visual position autoregressively along the codebook dimension, while masked diffusion refines all spatial positions in parallel.
Ablation studies confirm the contribution of both design choices.
Codebook-specific input embeddings improve next-code top-1 accuracy, with the largest gains on later codebooks.
The multi-codebook AR head captures inter-code dependencies and achieves the best results on all four generation benchmarks, with the most pronounced gains in counting, position, and attribute binding.

At the system level, the primary model is trained on 1,024 Huawei Ascend 910B NPUs and provides the main reported results.
A reference model, trained independently on NVIDIA GPUs with the same architecture and optimization recipe but a smaller training budget, is used to assess cross-platform consistency.
In addition, our proposed temporal-stability-guided sampling strategy reduces image generation latency by $4.45\times$ under the same evaluation protocol without degrading generation quality.
Overall, the results on omni-visual understanding, image generation, image editing, and physically grounded scientific-diagram generation demonstrate the potential of fully discrete diffusion language modeling to unify broad visual understanding and high-quality generation within a single framework.

Despite these advances, several limitations remain.
First, the multi-codebook autoregressive head introduces additional decoding overhead, and high-resolution generation produces long visual sequences.
Second, the current system covers image generation and editing, but has not yet been extended to video or 3D generation.
In future work, we plan to extend generation to videos and 3D assets, explore more tightly interleaved multimodal generation, investigate post-training with reinforcement learning, and scale up the backbone.

\FloatBarrier
\bibliography{references}

\clearpage
\appendix
\makeatletter\renewcommand{\@fb@secFB}{}\makeatother
\section{Special Tokens and Sequence Format Examples}
\label{app:tokens}

\subsection{Full model configuration}
\label{app:config}

Table~\ref{tab:config_full} lists the complete configuration of Intern Lumina U2 as stored in the checkpoint, together with the visual interface modules added in this work.

\begin{table}[!htbp]
  \centering
  \footnotesize
  \caption{Complete model configuration. Upper part: backbone settings. Lower
  part: visual interface modules.}
  \label{tab:config_full}
  \setlength{\tabcolsep}{3pt}
  \resizebox{\linewidth}{!}{%
  \begin{tabular}{llll}
    \toprule
    Field & Value & Field & Value \\
    \midrule
    Transformer layers & 20 & First dense layers & 1 \\
    Hidden size & 2048 & FFN size (dense) & 5120 \\
    Attention heads & 16 & KV heads & 4 \\
    Head dimension & 128 & RoPE dimension / fraction & 64 / 0.5 \\
    RoPE base & 600000 & Max position embeddings & 16384 \\
    RMSNorm / eps & yes / $10^{-6}$ & Activation & silu \\
    QKV bias / linear bias & no / no & QK norm & per-head RMSNorm \\
    Experts & 256 & Shared experts & 1 \\
    Expert FFN size & 512 & Experts per token & 8 \\
    Router score & sigmoid & Expert groups / selected & 8 / 4 \\
    Normalize top-$k$ probs & yes & Routed scaling factor & 2.5 \\
    Router dtype & fp32 & Expert selection bias & yes \\
    MoE implementation & fused & Attention implementation & sdpa / NPU fusion attn \\
    Vocabulary size & 157184 & Tied input/output embed.\ & no \\
    Padding token id & 156892 & Text mask token id & 156895 \\
    Parameter dtype & bfloat16 & Codebooks & 8 \\
    \midrule
    Visual input tables $E_k$ & $8\times(5010\times2048)$ & Input projection $W_{\mathrm{in}}$ & $16384\!\to\!2048$ \\
    Head condition projection $W_q$, $b_q$ & $2048\to2048$ & Position embeddings $s_k$ / $u_k$ & $8\times2048$ each \\
    Head depth / heads / MLP ratio & 6 / 8 / 4 & Head norm / activation & LayerNorm ($10^{-6}$) / GELU \\
    Head QK norm / adaLN & LN / 6-way per block & Output classifiers & $8\times(2048\!\to\!4096)$ \\
    Visual mask entry & 5000 & Text head & $2048\!\to\!157184$ \\
    Total / activated parameters & 16.9B / $\approx$1.4B & & \\
    \bottomrule
  \end{tabular}}
\end{table}

\subsection{Special tokens}
\label{app:special_tokens}

Table~\ref{tab:tokens} lists the corresponding special tokens.
Text-related tokens contains 157,184 tokens from LLaDA2.0 vocabulary and are embedded with the corresponding text embedding table.
Visual-related tokens are stored in each of the $\Kcb$ visual embedding tables, which have 5010 entries each: the first 4096 entries hold the codebook codes, and the entries after them hold the visual mask token and the delimiter tokens that mark the end of an image row, a video row or frame, and a 3D row or layer. Since a packed sequence is stored as an $[L,\Kcb]$ tensor with one column per codebook, each text or structural token is written into all $\Kcb$ columns, and the model reads it from the first column only.

\begin{table}[!htbp]
  \centering
  \small
  \caption{Special tokens. Ids in the lower block index the per-codebook
  visual embedding tables; all other ids index the text vocabulary.}
  \label{tab:tokens}
  \begin{tabular}{llr}
    \toprule
    Role & Token & Id \\
    \midrule
    Text mask (corruption target)      & \texttt{<|mask|>}       & 156895 \\
    Padding                            & \texttt{<|pad|>}        & 156892 \\
    Answer boundary                    & \texttt{<BOA>} / \texttt{<EOA>} & 157165 / 157166 \\
    Image boundary                     & \texttt{<BOI>} / \texttt{<EOI>} & 157167 / 157168 \\
    3D boundary                        & \texttt{<BO3D>} / \texttt{<EO3D>} & 157169 / 157170 \\
    Video boundary                     & \texttt{<BOV>} / \texttt{<EOV>} & 157171 / 157172 \\
    Media placeholders in text         & image / 3D / video context & 157160 / 157158 / 157161 \\
    Unconditional condition (CFG)      & \texttt{<uncondition>}  & 157156 \\
    Reasoning span                     & \texttt{<think>} / \texttt{</think>} & 157175 / 157176 \\
    User message wrapper (SFT only)    & user start / end        & 157155 / 157157 \\
    System wrapper (SFT only)          & system start / end      & 157153 / 157154 \\
    \midrule
    Visual mask entry                  & all $\Kcb$ codebooks        & 5000 \\
    Image row delimiter                & all $\Kcb$ codebooks    & 5001 \\
    3D row / layer delimiter           & all $\Kcb$ codebooks    & 5002 / 5003 \\
    Video row / frame delimiter        & all $\Kcb$ codebooks    & 5004 / 5005 \\
    \bottomrule
  \end{tabular}
\end{table}

\subsection{A complete packed sequence}
\label{app:seq_example}

Table~\ref{tab:seq_example} shows a complete image-understanding sample as an example for our constructed sequence.
The sample contains a three-token question and a three-token answer; for readability, the visual input is shown as a $2\times2$ grid instead of the $16\times16$ grid of an actual $256\times256$ image.
Before masking, the answer is padded with \texttt{<EOA>} tokens to the block length $B{=}32$.
The clean copy of the answer, by contrast, contains only the unpadded answer and reuses the position IDs of the corresponding noisy tokens.
Each position is also assigned a token type: type 1 denotes text conditions and structural tokens, type 2 media conditions, type 3 text answers, and type 4 visual answers.

\begin{table}[!htbp]
  \centering
  \footnotesize
  \caption{A packed image-understanding sample with per-position metadata. In this
  example, type 1 denotes text or structural tokens, type 2 a media-condition token,
  type 3 a text-answer token, and type 4 a visual-answer token. The clean
  user-provided condition and the assistant answer form one turn, and the clean copy of the answer belongs to the same turn. The \texttt{is\_clean} flag is 1 for
  clean tokens and 0 for noisy tokens, the block id is $-1$ for conditions and 0 for
  this answer block, and a clean copy reuses the position ids of its noisy counterpart.
  Here $v_{ij}$ is the visual token ($\Kcb$ codes) at grid row $i$, column $j$;
  $\lrcorner$ is the image-row delimiter (code 5001); $q_i$ and $a_i$ are question
  and answer tokens; and \texttt{M} is \texttt{<|mask|>}. Labels are $-100$ at all
  positions that do not participate in supervision; only masked answer positions
  contribute to the loss.}
  \label{tab:seq_example}
  \setlength{\tabcolsep}{3pt}
  \resizebox{\linewidth}{!}{%
  \begin{tabular}{rlccccl}
    \toprule
    idx & token & type & turn & is\_clean & block & position id \\
    \midrule
    \multicolumn{7}{l}{\emph{Turn 0: clean condition (media + question)}} \\
    0     & \texttt{<BOI>}                 & 1 & 0 & 1 & $-1$ & 0 \\
    1--2  & $v_{00}\; v_{01}$              & 2 & 0 & 1 & $-1$ & 1--2 \\
    3     & $\lrcorner$                    & 2 & 0 & 1 & $-1$ & 3 \\
    4--5  & $v_{10}\; v_{11}$              & 2 & 0 & 1 & $-1$ & 4--5 \\
    6     & $\lrcorner$                    & 2 & 0 & 1 & $-1$ & 6 \\
    7     & \texttt{<EOI>}                 & 1 & 0 & 1 & $-1$ & 7 \\
    8--10 & $q_1\; q_2\; q_3$              & 1 & 0 & 1 & $-1$ & 8--10 \\
    \midrule
    \multicolumn{7}{l}{\emph{Turn 0: noisy text answer (one block of $B{=}32$)}} \\
    11    & \texttt{<BOA>}                 & 1 & 0 & 0 & 0 & 11 \\
    12--14 & $a_1\; \texttt{M}\; a_3$ (label at idx 13 $= a_2$) & 3 & 0 & 0 & 0 & 12--14 \\
    15    & \texttt{<EOA>} or \texttt{M}   & 3 & 0 & 0 & 0 & 15 \\
    16--43 & 28 $\times$ (\texttt{<EOA>} or \texttt{M}); label \texttt{<EOA>} if masked & 3 & 0 & 0 & 0 & 16--43 \\
    \midrule
    \multicolumn{7}{l}{\emph{Turn 0: clean copy of the answer (natural length, no padding)}} \\
    44    & \texttt{<BOA>}                 & 1 & 0 & 1 & 0 & 11 \\
    45--47 & $a_1\; a_2\; a_3$             & 3 & 0 & 1 & 0 & 12--14 \\
    48    & \texttt{<EOA>}                 & 3 & 0 & 1 & 0 & 15 \\
    \bottomrule
  \end{tabular}}
\end{table}

Samples of the other tasks use the same format and differ only in the content of their turns. In a text-to-image sample, the single turn holds the prompt tokens (type 1, block $-1$), which are replaced by \texttt{<uncondition>} with probability 0.1, followed by \texttt{<BOA>} \texttt{<BOI>} grid \texttt{<EOI>} \texttt{<EOA>}, where the grid positions have type 4, the four boundary tokens have type 1, all answer tokens belong to block 0, and no clean copy is added. An edit sample likewise forms a single turn, in which the source image (type 2) and the instruction (type 1) are followed by the target image. A plain-text sample is a single turn that contains only an answer and its clean copy.

\section{Visual Token Layouts and Positions}
\label{app:layout}

For each visual modality, Table~\ref{tab:layout} lists the token grid produced by AToken, the order used to flatten the grid into a 1D sequence, and the delimiter tokens inserted into the sequence.
AToken applies its own 4D rotary position embedding within the tokenizer.
The backbone, in contrast, receives only the 1D sequence and applies partial RoPE (64 of 128 head dimensions, $\theta=6\times10^5$) to sequential position IDs.
Position IDs increase by one per token across the conditions and noisy answers in dialogue order; each clean copy reuses the position IDs of its noisy counterpart.

\begin{table}[!htbp]
  \centering
  \footnotesize
  \caption{Visual token layouts. Patch size is $(t,h,w)=(4,16,16)$; every grid
  position carries $\Kcb{=}8$ codes.}
  \label{tab:layout}
  \setlength{\tabcolsep}{3pt}
  \resizebox{\linewidth}{!}{%
  \begin{tabular}{llll}
    \toprule
    Modality & Grid & Flattening order & Delimiter tokens \\
    \midrule
    Image $H\times W$ & $(H/16)\times(W/16)$ & row-major $(y,x)$ & 5001 per row \\
    Video $T\times H\times W$ & $\lceil T/4\rceil\times(H/16)\times(W/16)$ & frame-major $(t,y,x)$ & 5004 per row, 5005 per frame \\
    3D asset & sparse voxels, coords $(x,y,z)$ & $(z,y,x)$ & 5002 per row, 5003 per layer \\
    \bottomrule
  \end{tabular}}
\end{table}

For images, a row delimiter is appended after each row of visual tokens.
Excluding the boundary tokens, a $256\times256$ image therefore occupies $16\times(16{+}1)=272$ positions, a $512\times512$ image $32\times33=1{,}056$ positions, and a $1024\times1024$ image $64\times65=4{,}160$ positions.
For videos and 3D assets, AToken returns the coordinates $(b,t,x,y,z)$ of non-empty positions only, and only these positions are included in the sequence.
A row delimiter is appended after each row containing at least one non-empty position, and a frame delimiter (for videos) or a layer delimiter (for 3D assets) is appended after each such frame or layer.
At the end of a frame or layer, the row delimiter precedes the frame or layer delimiter.

\section{Data Construction, Sources, and Audit Details}
\label{app:data}

This section provides the source categories, materialization rules, and validation steps of the training data.

\paragraph{Data sources by category.} Table~\ref{tab:data_sources} lists the public and private sources of each data category, together with their encoding resolution and the training stages in which they are used.

\begin{table}[!htbp]
  \centering
  \caption{Sources by data category, with the public and private sources of each category listed separately. Encoding resolution is the long side (images and editing) or the frame area budget (video); ``native'' means the original resolution clamped to a $[256,1024]$ long side. Stages are numbered 1--5 as in Section~\ref{sec:training}.}
  \label{tab:data_sources}
  \footnotesize
  \setlength{\tabcolsep}{3pt}
  \renewcommand{\arraystretch}{1.1}
  \begin{tabular}{>{\raggedright\arraybackslash}p{0.13\linewidth}>{\raggedright\arraybackslash}p{0.08\linewidth}>{\raggedright\arraybackslash}p{0.51\linewidth}>{\raggedright\arraybackslash}p{0.12\linewidth}>{\raggedright\arraybackslash}p{0.06\linewidth}}
    \toprule
    Category & Type & Sources & Encoding resolution & Stages \\
    \midrule
    Image understanding & Public & LLaVA-OneVision~\citep{Li2024_LLaVAOneVision} and LLaVA-OneVision-1.5~\citep{An2025_LLaVAOneVision15} instruction sets & 256 / 512 / native & 1--5 \\
     & Private & Image subsets of the InternVL~\citep{Wang2025_InternVL35} instruction data, covering captioning; document, chart, table, and equation data; OCR data built from web image--text pairs; grounding, counting, and GUI or web screenshot data; and structured-reasoning data and samples with \texttt{<think>} reasoning traces & & \\
    \midrule
    Video understanding & Public & Caption and QA sets over WebVid~\citep{Bain2021_WebVid}, Kinetics-400/600/700~\citep{Kay2017_Kinetics,Carreira2018_Kinetics600,Carreira2019_Kinetics700}, ActivityNet~\citep{Heilbron2015_ActivityNet}, YouCook2~\citep{Zhou2018_YouCook2}, LSMDC~\citep{Rohrbach2017_LSMDC}, TextVR~\citep{Wu2023_TextVR}, NTU RGB+D~\citep{Shahroudy2016_NTU}, NExT-QA~\citep{Xiao2021_NExTQA}, CLEVRER~\citep{Yi2020_CLEVRER}, STAR~\citep{Wu2021_STAR}, Perception Test~\citep{Patraucean2023_PerceptionTest}, EgoSchema~\citep{Mangalam2023_EgoSchema}, and MovieChat~\citep{Song2024_MovieChat} & 64 frames, $448^2$ area & 4--5 \\
     & Private & Video subsets of the InternVL instruction data; temporal-reasoning and counting sets & & \\
    \midrule
    3D understanding & Public & Objaverse~\citep{Deitke2023_Objaverse} assets with Cap3D-style captions~\citep{Luo2023_Cap3D}; GAPartNet~\citep{Geng2023_GAPartNet} part-level annotations; Sketchfab~\citep{Sketchfab2026} assets & sparse voxels & 4--5 \\
     & Private & -- & & \\
    \midrule
    Text-to-image & Public & Web image--text pairs (CC-style, $256^2$ only) & 256 / 512 / 1024 & 1--5 \\
     & Private & Curated real photographs with English and Chinese captions; synthetic images from FLUX, Ideogram, Recraft, Seedream, Dreamina, and Nano Banana; Chinese text-rendering, scientific-illustration, and design/typography sets; reasoning-prefixed (\texttt{t2i\_think}) subset & & \\
    \midrule
    Image editing & Public & OmniEdit~\citep{Wei2024_OmniEdit}, ImgEdit~\citep{Ye2025_ImgEdit}, SEED-Data-Edit~\citep{Ge2024_SEEDDataEdit}, and X2Edit~\citep{Ma2025_X2Edit} & 512 (some 1024) & 4--5 \\
     & Private & Single-turn attribute/color/material/style set; global-transformation set; precise compositional edits; multi-input samples; in-image text rendering including Chinese; GPT-4o-derived pairs & & \\
    \midrule
    Text and reasoning & Public & Nemotron post-training collections~\citep{Bercovich2025_LlamaNemotron} (chat, code, math, general); OpenMathInstruct-2~\citep{Toshniwal2024_OpenMathInstruct2}; Dolci-SFT~\citep{Olmo2025_Olmo3} & $\le$2,048 tokens & 1--5 \\
     & Private & The text-only portion of the InternVL instruction data; reasoning data with and without \texttt{<think>} traces & & \\
    \bottomrule
  \end{tabular}
\end{table}

\paragraph{Encoding of visual inputs.} All visual inputs are encoded by the frozen AToken. An image is converted to RGB with values scaled to $[-1,1]$ and resized with its aspect ratio preserved so that its long side equals the target resolution, after which both sides are rounded to multiples of 16 pixels with a minimum of 32 pixels. A video is represented by 64 frames chosen by segment-based sampling~\citep{Wang2016_TSN}, which draws one random frame from each of 64 equal-length segments, while clips with fewer frames keep all of their frames; the frames are resized so that their area fits a budget of $448^2$ pixels, and the frame count is padded to a multiple of the temporal patch size of 4. A 3D asset is voxelized at a resolution of $48^3$ and encoded into sparse voxels by the 3D branch of AToken, and assets without any occupied voxel are discarded. Most editing sources are encoded with a long side of 512 pixels, so that samples with two or three images fit within the packed sequence length, and part of the understanding data of the $1024^2$ stage is encoded at native resolution with the long side clamped to $[256,1024]$.

\paragraph{Data validation.} Before training, the annotations and sequence lengths of each data mixture are precomputed offline, and sources unchanged from a previous mixture are reused after their first sample is checked for consistency, while changed or new sources are processed again. The precomputed data pass four checks before use: (i) the numbers of sources and samples match the mixture definition; (ii) randomly drawn samples are non-empty and can be parsed; (iii) on a random subset, recomputed sequence lengths equal the stored ones; and (iv) a random subset can be converted into packed training sequences without error.

\paragraph{Text-to-image quality filtering.} As described in Section~\ref{sec:data}, the vision-language judge Qwen3-VL-30B-A3B-Instruct scores every image under the UniPercep protocol on four axes, each ranging from 0 to 100: aesthetics (composition, color and lighting, and artistic expression), image quality (sharpness, artifacts, and anatomical errors), structural richness, and prompt following (subject, count, attributes, scene, style, and text requirements). Under this protocol, the judge answers each axis with a two-letter code for one of four score bands (0--25, 26--50, 51--75, and 76--100), which avoids its bias toward particular digits, and the continuous score is the expectation over the bands weighted by the predicted code probabilities. Filtering then applies three rules in order: images with illegible or missing required text are discarded, images with a prompt-following score below 60 are discarded, and among the remaining images only those in the global top 70\% by aesthetic score, a threshold of about 74, are kept. Table~\ref{tab:t2i_funnel} reports the number of images removed by each rule for the two curated photograph sources.

\begin{table}[!htbp]
  \centering
  \caption{Quality filtering of the curated text-to-image sources (counts in millions).}
  \label{tab:t2i_funnel}
  \small
  \setlength{\tabcolsep}{5pt}
  \begin{tabular}{lrrrrr}
    \toprule
    Source & Scored & Text rule & Following $<60$ & Aesthetic top 70\% & Kept \\
    \midrule
    English & 22.69 & $-$4.33 & $-$4.75 & $-$3.09 & 10.51 (46\%) \\
    Chinese & 31.35 & $-$6.92 & $-$6.26 & $-$4.15 & 14.02 (45\%) \\
    \bottomrule
  \end{tabular}
\end{table}

\paragraph{Weighted sourcing.} Each source is weighted in a training mixture by its sampling multiplicity (Section~\ref{sec:data}), and the mixtures of the different stages in Table~\ref{tab:data_versions} assign different weights to the same source. Under-represented or harder capabilities receive higher weights, so that their share of training is not reduced by the larger sources.

\section{Ascend Adaptation Details}
\label{app:ascend}

Intern Lumina U2 is trained, evaluated, and served entirely on Ascend NPUs. Porting the code from GPUs requires changes in three respects, since the available operators, the interface semantics of the fused kernels, and the memory capacity of each NPU all differ from those of the GPU software stack, and the following paragraphs describe how each difference is handled.

\paragraph{Execution path.}
Every entry point imports \texttt{torch\_npu.contrib.transfer\_to\_npu} before any model code. This remaps \texttt{torch.cuda} onto the NPU runtime, so no \texttt{'cuda'} literal in the model files has to be rewritten. Each rank is given its own \texttt{HF\_MODULES\_CACHE} directory; otherwise the ranks sharing a node write the same dynamically imported module files at the same time. The AToken decoder ships with a \texttt{flash\_attn} dependency that does not exist on this platform, so its attention modules are rebuilt on the native kernels. That rebuild covers the image-decoding path; the understanding and text paths never reach that code.

\paragraph{Fused mixture-of-experts.}
We rebuild the fused MoE path on \texttt{torch\_npu.npu\_grouped\_matmul}, since the fused path of the GPU stack is a Triton kernel. The forward pass permutes tokens into expert order with \texttt{npu\_moe\_token\_permute}, then runs one grouped matrix multiplication against the merged first projection, \texttt{npu\_swiglu}, and a second grouped matrix multiplication against the second projection. The routing weights are applied as the result is unpermuted. The kernel provides no backward pass, so we write one as an explicit autograd function. It uses two further grouped matrix multiplications: one against the transposed weights for the input gradient, and one against the transposed activations, under a different grouping mode, for the weight gradient. An expert-parallel variant wraps the same kernel in all-to-all dispatch and combine.

\paragraph{Attention kernel semantics.}
Attention uses \texttt{torch\_npu.npu\_fusion\_attention} in the BNSD layout, with the TND layout and \texttt{actual\_seq\_qlen} available for variable-length batches. Two interface details change results silently, and both are handled explicitly. First, the mask polarity is inverted relative to the SDPA convention: the kernel reads \texttt{True} as masked, so the additive 4D mask is converted and negated rather than passed through. Second, the variable-length interface ignores \texttt{atten\_mask} and behaves as if no mask had been supplied, so it can express only full bidirectional attention or one template shared by every segment. The task-aware mask of Section~\ref{sec:seq_mask} gives each packed segment its own pattern, so packed training uses the dense path, where the mask reaches the kernel exactly as specified.

\paragraph{Memory.}
Per-NPU memory is the binding constraint of the training configuration, which makes activation allocation a first-order target rather than a detail absorbed by spare capacity. The allocator runs with \texttt{PYTORCH\_NPU\_ALLOC\_CONF=expandable\_segments:True} and \texttt{MULTI\_STREAM\_MEMORY\_REUSE=2}. Three further settings determine the usable ceiling. The data-parallel shard group size is 24 or 32 depending on the size of the job, which trades parameter and optimizer residency against communication volume. Gradient checkpointing is enabled on the backbone blocks. The cache-release interval is 20 steps rather than 100, which keeps fragmentation from accumulating across the long tail of packed sequence lengths. Together they lower peak memory and raise the longest sequence one NPU can hold, which is the quantity that sets training efficiency at fixed hardware.

\paragraph{Inference.}
A trained checkpoint stores the 256 experts of a layer as one stacked tensor. The platform provides a grouped matrix-multiplication kernel, so Ascend inference loads that layout directly and serves the training-time weights unchanged. The GPU path has no such kernel: it must first split the stacked tensor into per-expert weights, which costs a conversion pass and a second copy of the expert parameters.

\section{Additional Examples}
\label{app:samples}
\begin{figure}[t]
  \centering
  \includegraphics[width=\linewidth]{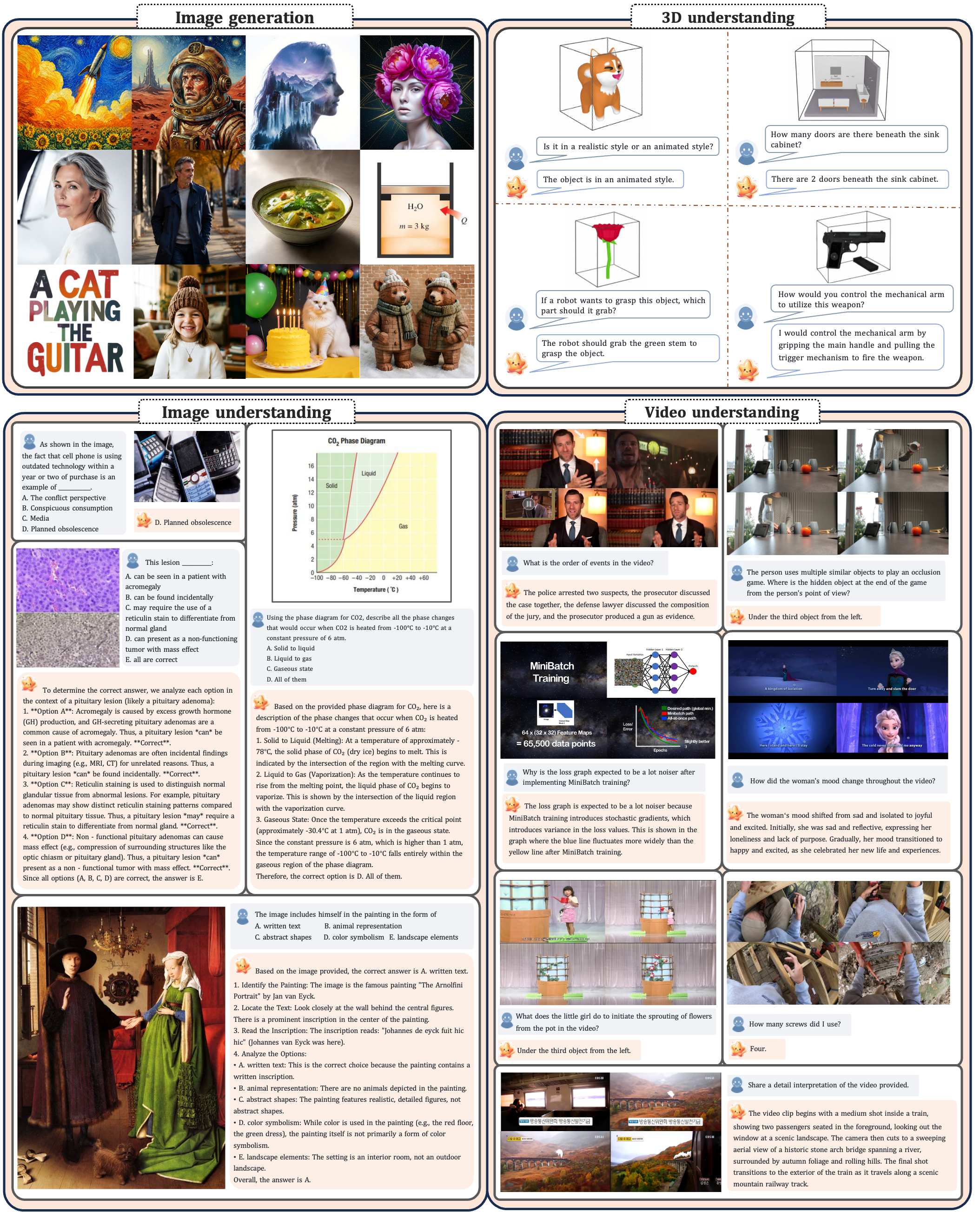}
  \caption{Qualitative examples for text-to-image generation and omni-visual
  understanding.}
  \label{fig:under_gen_qual}
\end{figure}

\begin{figure}[t]
  \centering
  \includegraphics[width=0.80\linewidth]{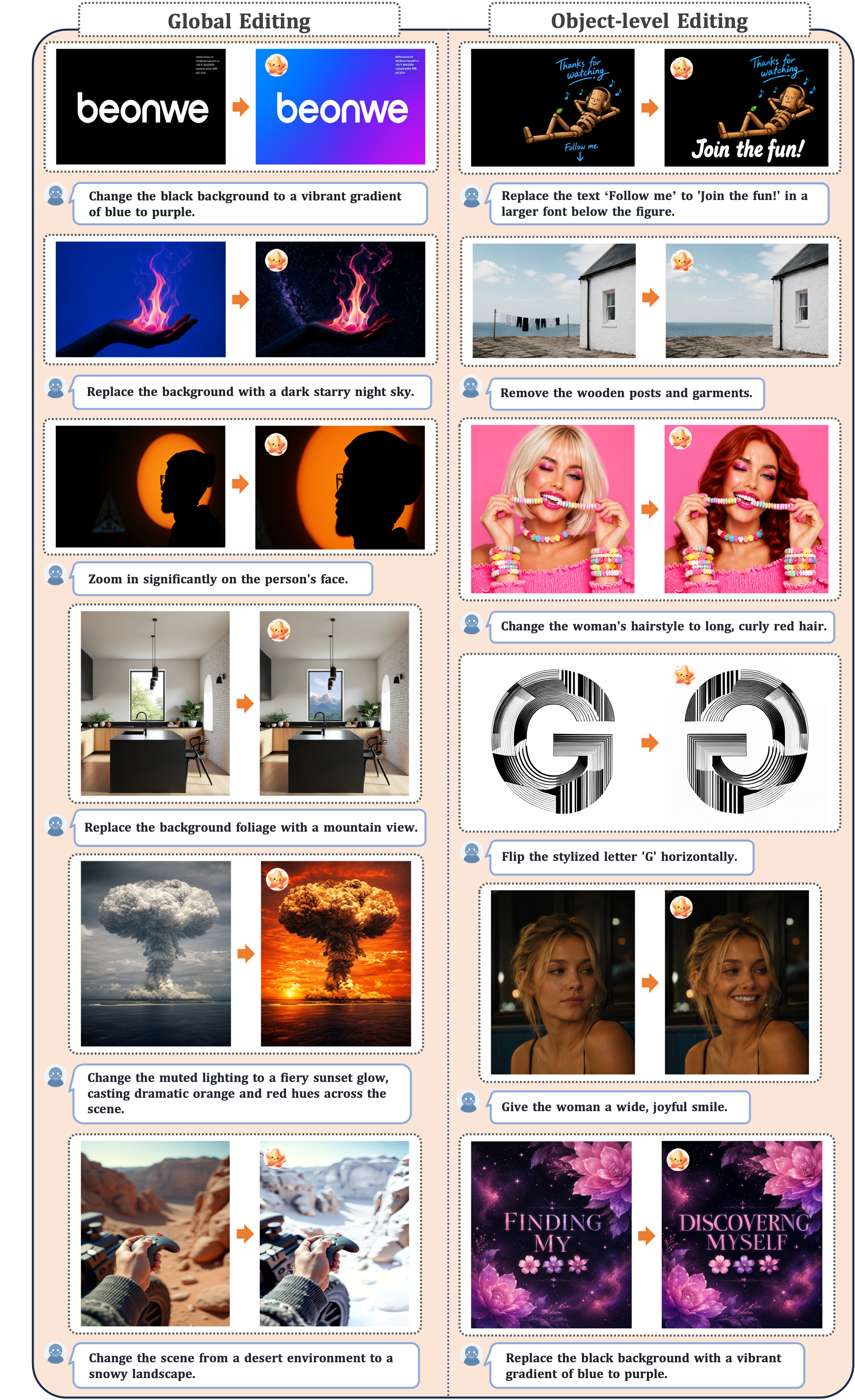}
  \caption{Qualitative examples for image editing.}
  \label{fig:edit_qual}
\end{figure}

Figures~\ref{fig:under_gen_qual} and~\ref{fig:edit_qual} provide qualitative
examples for the generation, omni-visual understanding, and editing tasks.

\end{document}